%% file: main.tex
\documentclass[10pt,twocolumn,letterpaper]{article}

\usepackage{cvpr}

\input{vcl-shortcuts.tex}

\input{my-shortcuts.tex}

\definecolor{somegray}{rgb}{0.5, 0.5, 0.5}
\newcommand{\darkgrayed}[1]{\textcolor{somegray}{#1}}
\makeatletter
\newcommand*\titleheader[1]{\gdef\@titleheader{#1}}
\AtBeginDocument{%
  \let\st@red@title\@title
  \def\@title{%
    \vskip-3.5em
    \bgroup\normalfont\large\centering\@titleheader\par\egroup
    \vskip1.5em\st@red@title}
}

\usepackage[pdfa,pagebackref=true,breaklinks=true,letterpaper=true,colorlinks,bookmarks=false]{hyperref}
\usepackage{balance}

\hypersetup{
  pdftitle={Events-to-Video: Bringing Modern Computer Vision to Event Cameras},
  pdfsubject={Computer Vision, Image Synthesis, Robotics},
  pdfauthor={Henri Rebecq, Ren\'e Ranftl, Vladlen Koltun, Davide Scaramuzza},
  pdfkeywords={Event cameras, Dynamic Vision Sensor, Image Synthesis}
}

\cvprfinalcopy %

\def\cvprPaperID{1595} %

\ifcvprfinal\fi

\makeatother

\titleheader{\darkgrayed{This paper has been accepted for publication at the\\ IEEE Conference on Computer Vision and Pattern Recognition (CVPR), Long Beach, 2019.
\copyright IEEE}}

\title{Events-to-Video: Bringing Modern Computer Vision to Event Cameras}

\begin{document}

\author{
Henri Rebecq \textsuperscript{\dag}
\and
Ren\'e Ranftl \textsuperscript{\ddag}
\and
Vladlen Koltun \textsuperscript{\ddag}
\and
Davide Scaramuzza \textsuperscript{\dag}
}

\maketitle

\nomarkerfootnote{\textsuperscript{\dag} Dept.~Informatics, Univ.~of Zurich and Dept.~Neuroinformatics, Univ.~of Zurich and ETH Zurich}
\nomarkerfootnote{\textsuperscript{\ddag} Intel Labs}

\begin{abstract}
Event cameras are novel sensors that report brightness changes in the form of asynchronous ``events'' instead of intensity frames.
They have significant advantages over conventional cameras: high temporal resolution, high dynamic range, and no motion blur.
Since the output of event cameras is fundamentally different from conventional cameras, it is commonly accepted that they require the development of specialized algorithms to accommodate the particular nature of events.
In this work, we take a different view and propose to apply existing, mature computer vision techniques to videos reconstructed from event data.
We propose a novel recurrent network to reconstruct videos from a stream of events, and train it on a large amount of simulated event data.
Our experiments show that our approach surpasses state-of-the-art reconstruction methods by a large margin ($>\!20\%$) in terms of image quality. We further apply off-the-shelf computer vision algorithms to videos
reconstructed from event data on tasks such as object classification and visual-inertial odometry, and show that this strategy consistently outperforms algorithms that were specifically designed for event data.
We believe that our approach opens the door to bringing the outstanding properties of event cameras to an entirely new range of tasks. %
A video of the experiments is available at \url{https://youtu.be/IdYrC4cUO0I}
\end{abstract}

\input{chapters/introduction.tex}

\input{chapters/related_work.tex}

\input{chapters/methodology.tex}

\input{chapters/exp_image_quality.tex}

\input{chapters/exp_classification.tex}

\input{chapters/exp_vio.tex}

\input{chapters/conclusion.tex}

\vspace{-1ex}
\section*{Acknowledgements}
\vspace{-1ex}
This work was supported by the the Swiss National Center of Competence Research Robotics (NCCR) and the SNSF-ERC Starting Grant.

\appendix

\input{chapters/supplement.tex}

\clearpage

{\small
\balance
\bibliographystyle{ieee}
\bibliography{all}
}

\end{document}

%% file: vcl-shortcuts.tex
\usepackage{times}
\usepackage{epsfig}
\usepackage{graphicx}
\usepackage{float}
\usepackage{wrapfig}
\usepackage{amsmath,amssymb,amsthm}
\usepackage{bm,xspace}
\usepackage{comment}
\usepackage{verbatim}
\usepackage{multirow}
\usepackage{balance}
\usepackage{url}
\usepackage{booktabs}
\usepackage{etoolbox,siunitx}
\usepackage{calc}
\usepackage{pifont,hologo}
\usepackage[usenames, dvipsnames]{color}
\usepackage{nicefrac}

\newcommand{\ra}[1]{\renewcommand{\arraystretch}{#1}}

\usepackage[super]{nth}
\usepackage{nicefrac}
\robustify\bfseries

\def\ee{\mathbf{e}}

\def\uu{\mathbf{u}}

\def\xx{\mathbf{x}}

\def\EE{\mathbf{E}}

\def\II{\mathbf{I}}

\def\iI{\mathcal{I}}

\def\lL{\mathcal{L}}

\def\eps{\varepsilon}

\DeclareMathSymbol{@}{\mathord}{letters}{"3B}

\newcommand\timess{\mathbin{\!\times\!}}

\newcommand\revised[1]{\textcolor{black}{#1}}

\newcommand\mypara[1]{\vspace{1mm}\noindent\textbf{#1}}

\def\latex/{\LaTeX}
\def\bibtex/{\hologo{BibTeX}}

%% file: my-shortcuts.tex
\newcommand\nomarkerfootnote[1]{%
  \begingroup
  \renewcommand\thefootnote{}\footnote{#1}%
  \addtocounter{footnote}{-1}%
  \endgroup
}

\def\Event{\ee}
\def\EventSequence{\eps}
\def\SensorWidth{W}
\def\SensorHeight{H}

\def\VoxelGrid{\EE}
\def\NumEvents{N}
\def\NumBins{B}
\def\Loss{\lL}
\def\Image{\iI}
\def\EventTensor{\mathbf{\EE}}
\def\Reconstructed{\hat{\iI}}
\def\ImageSequence{\II^S}
\def\EventTensorSequence{\EventTensor^S}
\def\Brightness{L}
\def\Pixel{\uu}

%% file: chapters/introduction.tex
\vspace{-0.3cm}
\section{Introduction}

Event cameras are bio-inspired vision sensors that work radically differently from conventional cameras. Instead of capturing intensity images at a fixed rate, event cameras measure \emph{changes} of intensity asynchronously at the time they occur. This results in a stream of \emph{events}, which encode the time, location, and polarity (sign) of brightness changes (Fig.~\ref{fig:principle_operation}). Event cameras such as the Dynamic Vision Sensor (DVS)~\cite{Lichtsteiner08ssc} possess outstanding properties when compared to conventional cameras. They have a very high dynamic range (\SI{140}{\decibel} versus \SI{60}{\decibel}), do not suffer from motion blur, and provide measurements with a latency as low as one microsecond. Event cameras thus provide
a viable alternative, or complementary, sensor in conditions that are challenging for conventional cameras.

\begin{figure}[t]
\centering
\setlength{\tabcolsep}{-0.1cm}
\begin{tabular}{ccc}
  \includegraphics[trim={1.5cm 1.2cm 1.5cm 1.5cm},clip,height=2.65cm]{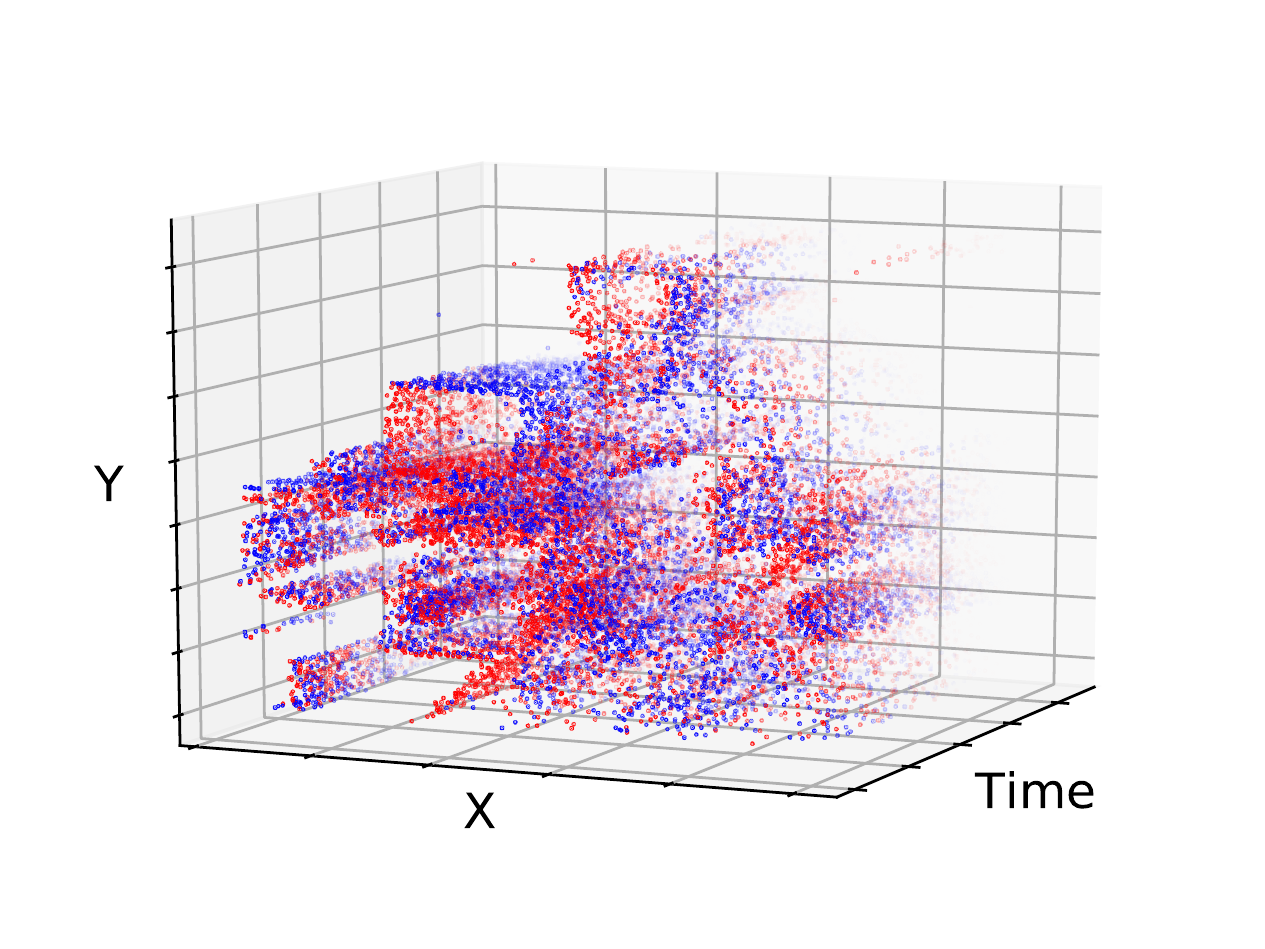} &
  \includegraphics[width=1.6cm,trim={0 -2.1cm 0 0cm},clip]{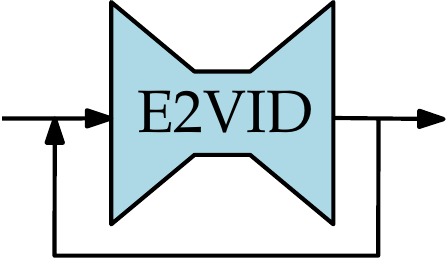}\hspace{0.3cm} &
  \includegraphics[height=3.0cm]{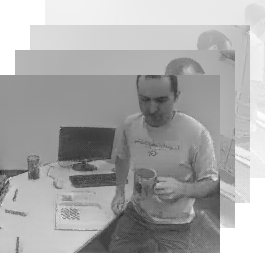}\\[-0.2mm]
  \multicolumn{3}{c}{\hspace{0mm}\includegraphics[width=0.85\linewidth]{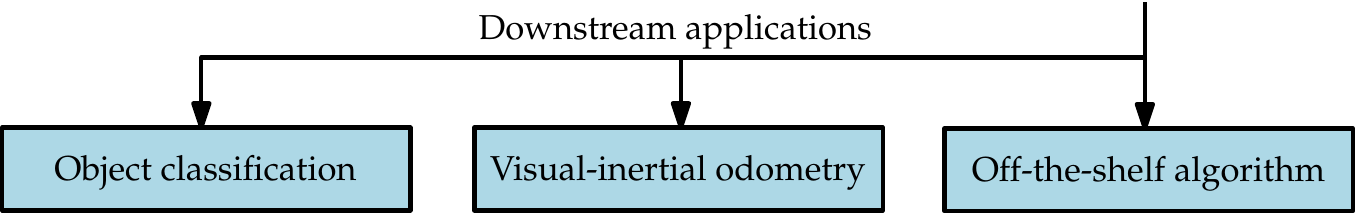}}
\end{tabular}
\vspace{1.0ex}
\caption{Our network converts a spatio-temporal stream of events (left) into a high-quality video (right).
This enables direct application of off-the-shelf computer vision algorithms such as object classification (Section~\ref{sec:exp_classification}) and visual-inertial odometry (Section~\ref{sec:exp_vio}), yielding state-of-the-art results with event data in each case.
This figure shows an actual output sample from our method, operating on a real event sequence from a publicly available dataset \cite{Mueggler17ijrr}.
\vspace{-2mm}
\label{fig:eye_catcher}
}
\end{figure}

However, since the output of an event camera is an asynchronous stream of events (a representation
that is fundamentally different from natural images),
existing computer vision techniques cannot be directly applied to this data. As a consequence, custom algorithms need to be specifically tailored to leverage event data. Such specialized algorithms have demonstrated impressive performance in applications ranging from low-level vision tasks, such as visual
odometry \cite{Kim16eccv, Rebecq17ral, Zhu17cvpr, Rebecq17bmvc, Rosinol18ral}, feature tracking \cite{Kueng16iros,Zhu17icra,Gehrig18eccv} and optical flow \cite{Benosman14tnnls,Bardow16cvpr,Stoffregen17acra,Zhu18rss,Stoffregen19arxiv}, to high-level tasks such as object classification \cite{PerezCarrasco13pami,Lagorce16pami,Sironi18cvpr} and gesture recognition \cite{Amir17cvpr}.

While some works \cite{Conradt09iscas,Cook11ijcnn,Benosman14tnnls,Mueggler14iros,Kim16eccv,Gallego17pami} focused on exploiting the low latency of the sensor by processing the data event-by-event, significant progress has been made by mapping a set of events into an image-like 2D representation prior to processing. Examples are the integration of events on the image plane \cite{Rebecq17ral,Liu18bmvc,Maqueda18cvpr} as well as time surfaces \cite{Lagorce16pami,Sironi18cvpr,Zhu18rss,Zhou18eccv,Zhu18rss}.
However, neither event images nor time surfaces are natural images, meaning that much of the existing computer vision toolbox cannot be applied effectively. Most importantly, deep networks trained on real image data cannot be directly transferred to these representations.

In this paper, we propose to establish a bridge between vision \revised{with event cameras} and conventional computer vision.
Specifically, we learn how to reconstruct natural videos from a stream of events (``events-to-video''), \ie we learn a mapping between a stream of events and a stream of images (Fig.~\ref{fig:eye_catcher}).
This allows us to apply off-the-shelf computer vision techniques to event cameras.

Our work differs from previous image reconstruction approaches \cite{Bardow16cvpr,Munda18ijcv,Scheerlinck18accv} in two essential ways.
First, instead of embedding handcrafted smoothness priors into our reconstruction framework, we directly learn video reconstruction from events using a large amount of simulated event data. Second, instead of focusing mainly on the quality of the reconstructions, we build our approach with the goal of applying standard computer vision techniques to the reconstructions.
To this end, we encourage the reconstructed images to share the statistics of natural images through a perceptual loss that operates on mid-level image features.

To further validate the quality of our approach, we use our reconstructions to solve two popular problems \revised{with event cameras}: (i) object classification from a stream of events, and (ii) visual-inertial odometry. We apply off-the-shelf computer vision algorithms that were built to process conventional images to the reconstructed videos for both tasks. We show that this strategy outperforms state-of-the-art methods that had been specifically designed for event data.

In summary, our contributions are:
\begin{itemize}
  \item A novel recurrent network architecture to reconstruct a video from a stream of events, which outperforms the state-of-the-art in terms of image quality by a large margin.
  \vspace{-1ex}
  \item We establish that the network can be trained from simulated event data and generalizes remarkably well to real events.
  \vspace{-1ex}
  \item The application of our method to two problems \revised{with event cameras}: object classification and visual-inertial odometry \revised{from event data}. Our method outperforms state-of-the-art algorithms designed specifically for event data in both applications.
\end{itemize}

We believe that the most alluring characteristic of our method is that it acts as a bridge between conventional cameras and event cameras, thus bringing the main stream of computer vision research to event cameras: mature algorithms, modern deep network architectures, and weights pretrained from large natural image datasets. %
We believe that our work will open the door to leveraging the benefits of event cameras -- high temporal resolution, high dynamic range (Fig.~\ref{fig:qualitative_night}), and no motion blur (Fig.~\ref{fig:vio_motion_blur_frames}) -- to a broader array of applications.

%% file: chapters/related_work.tex
\section{Related Work}
\label{sec:related_work}

Events-to-video reconstruction is a popular topic \revised{in the event camera literature}.
Early approaches did not reconstruct videos, but focused on the reconstruction of a single image from a large set of events collected by an event camera moving through a static scene. These works exploit the fact that every event provides one equation relating the intensity gradient and optic flow through brightness constancy \cite{Gehrig18eccv}.
Cook \etal~\cite{Cook11ijcnn} used bio-inspired, interconnected networks to simultaneously recover intensity images, optic flow, and angular velocity from an event camera performing small rotations.
Kim \etal~\cite{Kim14bmvc} developed an Extended Kalman Filter to reconstruct a 2D panoramic gradient image (later upgraded to a full intensity frame by 2D Poisson integration) from a rotating event camera, and later extended it to a 3D scene and 6 degrees-of-freedom (6DOF) camera motion \cite{Kim16eccv} (albeit in a static scene only).
Bardow \etal~\cite{Bardow16cvpr} proposed to estimate optic flow and intensity \emph{simultaneously} from sliding windows of events through a variational energy minimization framework.  %
They showed the first video reconstruction framework from events that is applicable to dynamic scenes. %
However, their energy minimization framework requires multiple hand-crafted regularizers, which can result in severe loss of detail in the reconstructions. %

Recently, methods based on direct event integration have emerged. These approaches do not rely on any assumption about the scene structure or motion dynamics, and can naturally reconstruct videos at arbitrarily high framerates.
Munda \etal \cite{Munda18ijcv} cast intensity reconstruction as an energy minimization problem defined on a manifold induced by the event timestamps. They combined direct event integration with total variation regularization and achieved real-time performance on the GPU.
Scheerlinck \etal \cite{Scheerlinck18accv} proposed to filter the events with a high-pass filter prior to integration, and demonstrated video reconstruction results that are qualitatively comparable with \cite{Munda18ijcv} while being computationally more efficient.
While these approaches currently define the state-of-the-art, both suffer from artifacts which are inherent to direct event integration. The reconstructions suffer from ``bleeding edges'' caused by the fact that the contrast threshold (the minimum brightness change of a pixel to trigger an event) is neither constant nor uniform across the image plane.
Additionally, pure integration of the events can in principle only recover intensity up to an unknown initial image $\Image_0$, which causes ``ghosting'' effects where the trace of the initial image remains visible in the reconstructed images.

Barua \etal \cite{Barua16wacv} proposed a learning-based approach to reconstruct intensity images from events.
They used \mbox{K-SVD} \cite{Aharon06tsp} on simulated data to learn a dictionary that maps small patches of integrated events to an image gradient and used Poisson integration to recover the intensity image.
In contrast to \cite{Barua16wacv}, we do not reconstruct individual intensity images from small windows of events, but instead reconstruct a temporally consistent video from a long stream of events (several seconds) using a recurrent network. Instead of mapping event patches to a dictionary of image gradients, we learn pixel-wise intensity estimation directly. %

Despite the body of work on events-to-video reconstruction, further downstream vision applications based on the reconstructions have, to the best of our knowledge, never been demonstrated prior to our work.

%% file: chapters/methodology.tex
\section{Video Reconstruction Approach}
\label{sec:methodology}

\global\long\def\introPlotHeight{2.55cm}
\begin{figure}[t]
\centering
\setlength{\tabcolsep}{0.06cm}
\begin{tabular}{c}
  \includegraphics[height=3.0cm]{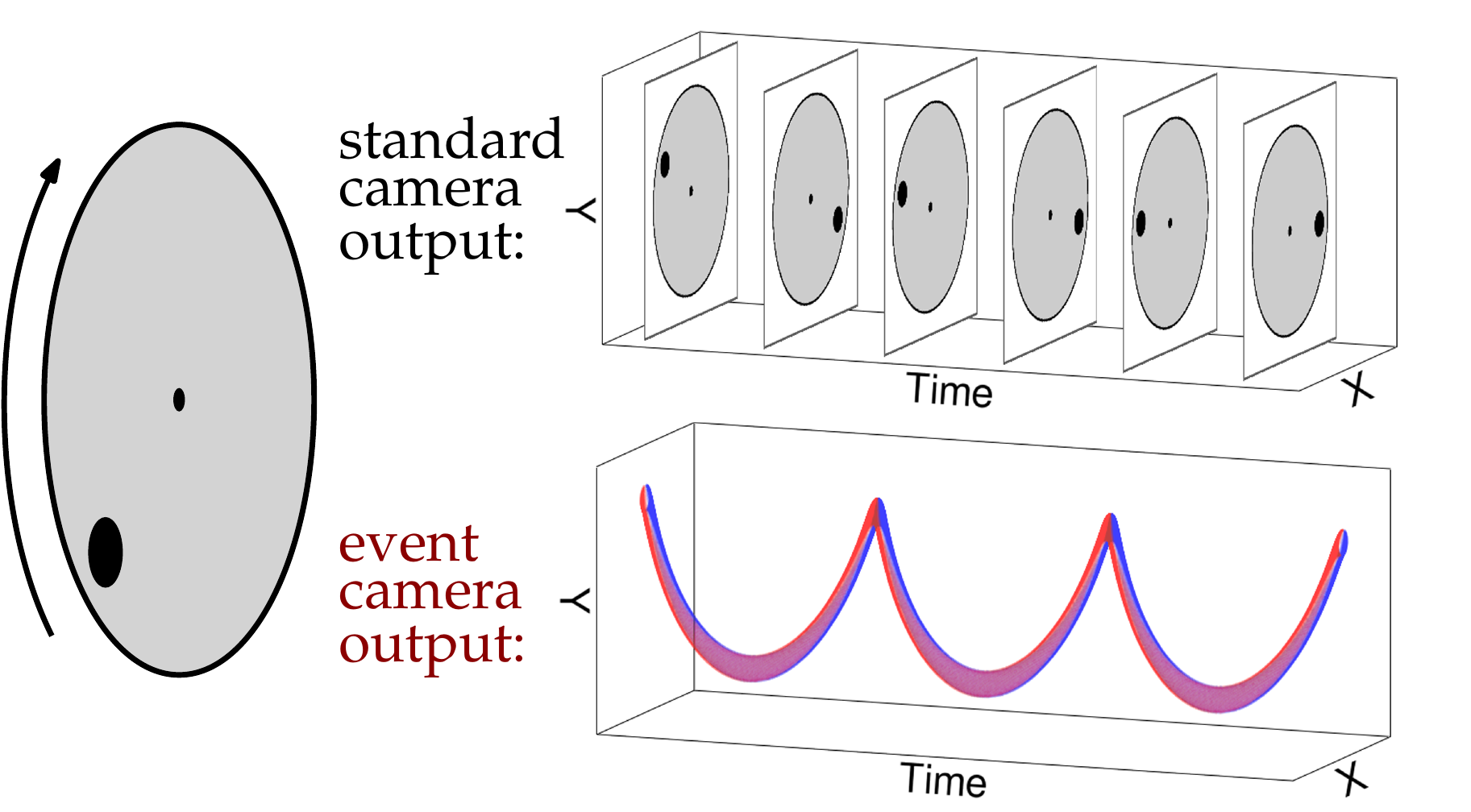}
\end{tabular}
\caption{Comparison of the output of a conventional camera and an event camera looking at a black disk on a rotating circle.
While a conventional camera captures frames at a fixed rate, an event camera transmits the brightness changes continuously in the form of a spiral of events in space-time (red: positive events, blue: negative events). Figure inspired by \cite{Mueggler14iros}.
}
\label{fig:principle_operation}
\end{figure}

An event camera consists of independent pixels that respond to changes in the spatio-temporal brightness signal $\Brightness(\xx,t)$\footnote{Event cameras respond in fact to logarithmic brightness changes, i.e. $\Brightness=\log E$ where $E$ is the \emph{irradiance}.} and transmit the changes in the form of a stream of asynchronous events~(Fig.~\ref{fig:principle_operation}).
For an ideal sensor, an event $\Event_i=(\Pixel_i,t_i,p_i)$ is triggered at pixel $\Pixel_i=(x_i,y_i)^T$ and time $t_i$ when the brightness change since the last event at the pixel reaches a threshold $\pm C$, which can be fixed by the user.
However, $C$ is in reality neither constant nor uniform across the image plane.
Instead, it strongly varies depending on various factors, such as the sign of the brightness change~\cite{Gallego17pami}, the event rate (because of limited pixel bandwidth)~\cite{Brandli14iscas}, and the temperature~\cite{Lichtsteiner08ssc}. Consequently, events cannot by directly integrated to recover accurate intensity images in practice.

\subsection{Overview}

\begin{figure}
	\centering
	\begin{tabular}{c}
    \includegraphics[height=3.0cm]{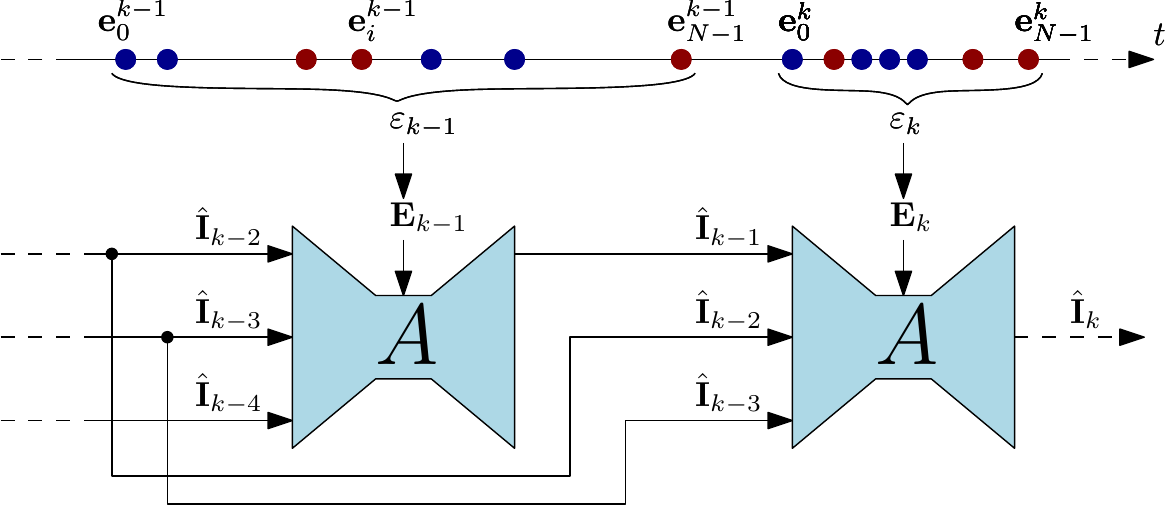}
    \end{tabular}
\caption{Overview of our approach. The event stream (depicted as red/blue dots on the time axis) is split into windows $\EventSequence_k$ with $\NumEvents$ events in each. Each window is converted into a 3D event tensor $\EventTensor_k$ and passed through the network together with the last $K$ reconstructed images to generate a new image reconstruction $\Reconstructed_k$. In this example, $K=3$ and $\NumEvents=7$.
}
\label{fig:overview}
\end{figure}

Our goal is to translate a continuous stream of events
into a sequence of images $\lbrace \Reconstructed_k \rbrace$, where $\Reconstructed_k \in \left[0,1\right]^{W\timess H}$.
To achieve this, we partition the incoming stream of events into sequential (non-overlapping) spatio-temporal windows of events $\EventSequence_k=\left\lbrace \Event_i \right\rbrace$, for $i \in \left[ 0,\NumEvents-1 \right]$,
each containing a fixed number $\NumEvents$ of events.
For each new event sequence $\EventSequence_k$, we generate a new image $\Reconstructed_k$ by fusing the $K$ previous reconstructed images $\lbrace \Reconstructed_{k-K}, ..., \Reconstructed_{k-1} \rbrace$ with the new events $\EventSequence_k$ (see Fig.~\ref{fig:overview}).
The reconstruction function is implemented by a recurrent convolutional neural network. %
We train the network in supervised fashion, using a large quantity of simulated event sequences with corresponding ground-truth images.
Because we process windows with a constant number of events, the output framerate is proportional to the event rate, making our approach fully data-driven. %
While our method introduces some latency due to processing events in windows, it nonetheless captures the major advantages of event cameras: our reconstructions have a high dynamic range (Fig.~\ref{fig:qualitative_night}) and are free of motion blur, even at high speeds~(Fig.~\ref{fig:vio_motion_blur_frames}).

\subsection{Event Representation}

In order to be able to process the event stream using a CNN, we need to convert $\EventSequence_k$ into a fixed-size tensor representation $\EventTensor_k$.
A natural choice is to encode the events in a spatio-temporal voxel grid~\cite{Zhu18eccvw}.
The duration ${\Delta T=t^k_{N-1}-t^k_0}$ spanned by the events in $\EventSequence_k$ is discretized into $\NumBins$ temporal bins.
Every event distributes its polarity $p_i$ to the two closest spatio-temporal voxels as follows:
\begin{equation}
  \VoxelGrid(x_l,y_m,t_n) = \sum_{\substack{x_i = x_l\\y_i = y_m}}{p_i \max(0,1-|t_n-t^{*}_i|)},
\end{equation}
where
$t^{*}_i \triangleq \frac{\NumBins-1}{\Delta T}(t_i-t_0)$ is the normalized event timestamp.
We use $\NumEvents=25@000$ events per window and $\NumBins=10$ temporal bins, unless specified otherwise.

\subsection{Training Data}
\label{sec:training_data_generation}

Our network requires training data, \ie a large amount of event sequences with corresponding ground-truth image sequences.
Formally, if we let $\EventTensorSequence=\left\lbrace\VoxelGrid_0, ..., \VoxelGrid_{T-1} \right\rbrace$ be a sequence of event tensors, and $\ImageSequence=\left\lbrace\Image_0, ..., \Image_{T-1}\right\rbrace$ be the corresponding sequence of images, we need to generate a large dataset of mappings $\left\lbrace \EventTensorSequence \leftrightarrow \ImageSequence \right\rbrace$.
However, there exists no such large-scale dataset with event data and corresponding ground-truth images.
Furthermore, images acquired by a conventional camera would provide poor ground truth in scenarios where event cameras excel, namely high dynamic range and high-speed scenes.
For these reasons, we propose to train the network on synthetic event data, and show subsequently that our network generalizes to real event data in Section~\ref{sec:image_quality}.

We use the event simulator ESIM \cite{Rebecq18corl}, which allows simulating a large amount of event data reliably.
ESIM renders images along the camera trajectory at high framerate, and interpolates the brightness signal at each pixel to approximate the continuous intensity signal needed to simulate an event camera.
Consequently, ground-truth images $\Image$ are readily available.
We map MS-COCO images~\cite{TsungYi14eccv} to a 3D plane, and simulate the events triggered by random camera motion within this simple 3D scene.
Examples of generated synthetic event sequences are presented in the appendix. %
We enrich the training data by simulating a different set of positive and negative contrast thresholds for each simulated scene (sampled according to a normal distribution with mean $0.18$ and standard deviation $0.03$; these parameters were chosen based on empirical data).
This prevents the network from learning to simply integrate events, which would work on noise-free, simulated data, but would generalize poorly to real event data (for which the assumption of a fixed contrast threshold does not hold).
The camera sensor size is set to $240 \times 180$ pixels (to match the resolution of the DAVIS240C sensor \cite{Brandli14ssc} used in our evaluation). %
Using \mbox{MS-COCO} images allows capturing a much larger variety of scenes than is available in any existing event camera dataset.
We generate $1@300$ sequences of 2 seconds each, which results in approximately 45 minutes of simulated event data. %
Note that the simulated sequences contain only globally homographic motion (\ie there is no independent motion in the simulated sequences).
Nevertheless, our network generalizes surprisingly well to scenes with arbitrary motions, as will be shown in Section~\ref{sec:image_quality}.

\subsection{Network Architecture and Training}
The main module of our recurrent network is a UNet \cite{Ronneberger15icmicci} architecture similar to the one introduced by Zhu \etal \cite{Zhu18rss} in the context of optical flow estimation.
The input tensor (obtained by concatenating $\EventTensor_k,\Reconstructed_{k-K}, ..., \Reconstructed_{k-1}$), of size $(\NumBins + K) \timess \SensorHeight \timess \SensorWidth$, is passed through 4 strided convolution layers (the number of output channels doubling each time), followed by two residual blocks~\cite{He16cvpr} and four upsampling transposed convolution layers.
The resulting activation is convolved depthwise to obtain a final image reconstruction.
Following \cite{Zhu18rss}, we use skip connections between symmetric convolution layers.
\revised{Additional details of the architecture are provided in the appendix.}
On top of this basic module (labeled ``A'' in Fig.~\ref{fig:overview}), we introduce a recurrent connection to propagate intensity information forward in time; in other words, the network does not need to reconstruct a new image from scratch at every time step, but only to incrementally update the previous reconstructions using the new sequence of events.
During training we unroll the network for $L$ steps.
At test time, the preceding $K$ reconstructed images are fed into the network (Fig.~\ref{fig:overview}).
We found that $L=8$ and $K=3$ provide a good trade-off between reconstruction quality and training time.

\mypara{Loss:}
We use the calibrated perceptual loss (LPIPS) \cite{Zhang18cvpr}, which passes the reconstructed image and the target image through a VGG network \cite{Simonyan14arxiv} trained on ImageNet~\cite{Russakovsky15ijcv}, and averages the distances between VGG features computed across multiple layers.
By minimizing LPIPS, our network effectively learns to endow the reconstructed images with natural statistics (\ie with features close to those of natural images).
The total loss $\Loss_k$ is computed as $\Loss_k=\sum_{l=0}^{L}{d_{L}(\Reconstructed_{k-l}, \Image_{k-l})}$,
where $d_{L}$ denotes the LPIPS distance.

\mypara{Training Procedure:}
We split the synthetic sequences into $1@270$ training sequences and $30$ validation sequences, and implement our network using PyTorch \cite{Paszke17nipsw}.
We use ADAM~\cite{Kingma15iclr} with an initial learning rate of $0.0001$, subsequently decayed by a factor of $0.9$ every $10$ epochs.
We use a batch size of $16$ and train for $40$ epochs.%

%% file: chapters/exp_image_quality.tex
\begin{flushright}
•
\end{flushright}\section{Evaluation}
\label{sec:image_quality}

\setlength{\tabcolsep}{0.3ex} %
\global\long\def\heightplot{2.1cm} %
\global\long\def\widthplot{2.805cm} %
\global\long\def\vspacecols{0.15ex} %
\begin{figure*}[t]
	\centering
    \begin{tabular}{cccccc}
    \includegraphics[width=\widthplot,height=\heightplot]{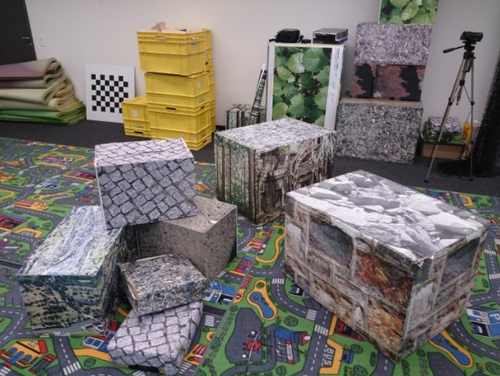}
    & \includegraphics[width=\widthplot,height=\heightplot]{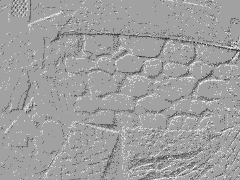}
    & \includegraphics[width=\widthplot,height=\heightplot]{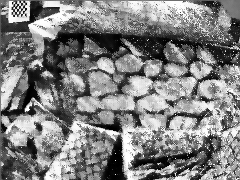}
    & \includegraphics[width=\widthplot,height=\heightplot]{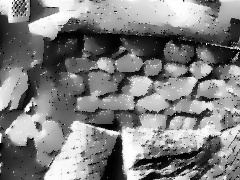}
    & \includegraphics[width=\widthplot,height=\heightplot]{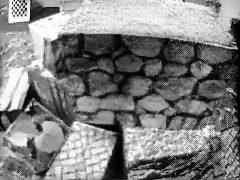}
    & \includegraphics[width=\widthplot,height=\heightplot]{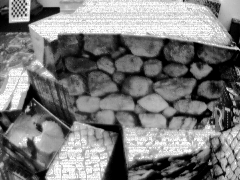}\\[\vspacecols]

    \includegraphics[width=\widthplot,height=\heightplot]{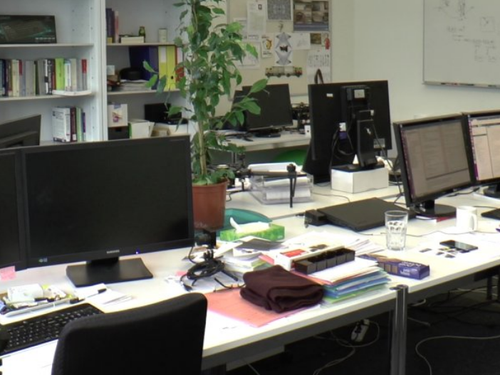}
    & \includegraphics[width=\widthplot,height=\heightplot]{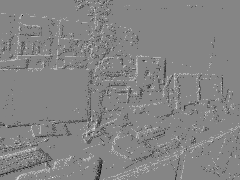}
    & \includegraphics[width=\widthplot,height=\heightplot]{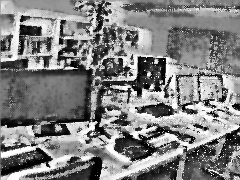}
    & \includegraphics[width=\widthplot,height=\heightplot]{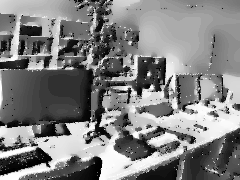}
    & \includegraphics[width=\widthplot,height=\heightplot]{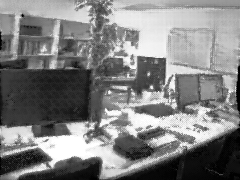}
    & \includegraphics[width=\widthplot,height=\heightplot]{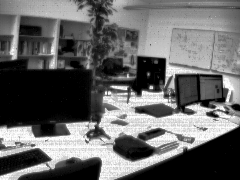}\\[\vspacecols]
    \includegraphics[width=\widthplot,height=\heightplot]{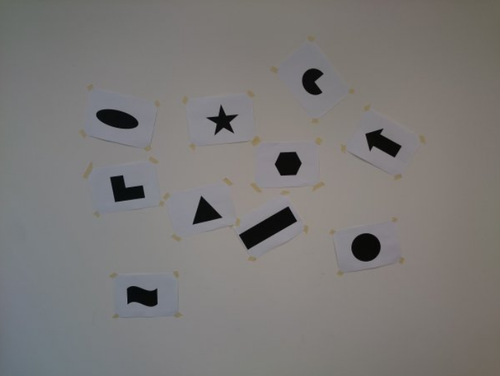}
    & \includegraphics[width=\widthplot,height=\heightplot]{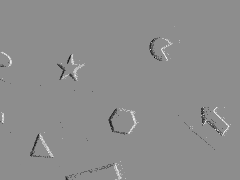}
    & \includegraphics[width=\widthplot,height=\heightplot]{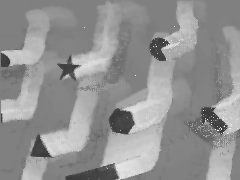}
    & \includegraphics[width=\widthplot,height=\heightplot]{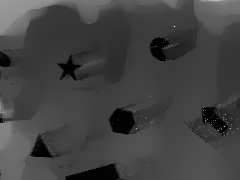}
    & \includegraphics[width=\widthplot,height=\heightplot]{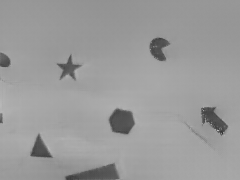}
    & \includegraphics[width=\widthplot,height=\heightplot]{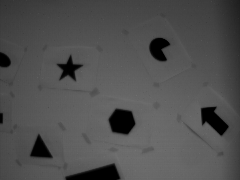}\\[\vspacecols]
    (a) Scene overview & (b) Events & (c) HF & (d) MR & (e) Ours & (f) Ground truth\\
    \end{tabular}
\vspace{0.5ex}
\caption{Comparison of our method with MR and HF on sequences from \cite{Mueggler17ijrr}. %
Our network reconstruct fine details well (textures in the first row) compared to the competing methods, while avoiding their artifacts (e.g. the ``bleeding edges'' in the third row).
}
\label{fig:comp_event_camera_dataset}
\end{figure*}

In this section, we present both quantitative and qualitative results on the fidelity of our reconstructions, and compare to recent methods \cite{Bardow16cvpr,Munda18ijcv,Scheerlinck18accv}. We focus our evaluation on real event data. An evaluation on synthetic data can be found in supplementary material.

We use event sequences from the Event Camera Dataset~\cite{Mueggler17ijrr}. These sequences were recorded using a DAVIS240C sensor \cite{Brandli14ssc} moving in various environments.
It contains events as well as ground-truth grayscale frames at a rate of \SI{20}{\Hz}.
We remove the redundant sequences (\eg that were captured in the same scene) and those for which the frame quality is poor, leaving seven sequences in total that amount to $1@670$ ground-truth frames.
For each sequence, we reconstruct a video from the events with our method and each baseline.
For each ground truth frame, we query the reconstructed image with the closest timestamp to the ground truth (tolerance of $\pm \SI{1}{\ms}$).

Each reconstruction is then compared to the corresponding ground-truth frame according to several quality metrics.
We equalize the histograms of every ground-truth frame and reconstructed frame prior to computing the error metrics (this way the intensity values lie in the same intensity range and are thus comparable).
Note that the camera speed gradually increases in each sequence, leading to significant motion blur on the ground-truth frames towards the end of the sequences; we therefore exclude these fast sections in our quantitative evaluation.
We also omit the first few seconds from each sequence, which leaves enough time for the baseline methods that are based on event integration to converge.
Note that this works in favor of the baselines, as our method converges almost immediately (the initialization phase is analyzed in the supplementary material).

We compare our approach against several state-of-the-art methods: \cite{Bardow16cvpr} (which we denote as SOFIE for ``Simultaneous Optic Flow and Intensity Estimation''), \cite{Scheerlinck18accv} (HF for ``High-pass Filter''), and \cite{Munda18ijcv} (MR for ``Manifold Regularization'').
For HF and MR, we used the code that was provided by the authors and manually tuned the parameters on the evaluated sequences to get the best results possible. %
For HF, we also applied a bilateral filter to the reconstructed images (with filter size $d=5$ and $\sigma=25$) in order to remove high-frequency noise, which improves the results of HF in all metrics.
For SOFIE, we report qualitative results instead of quantitative results since we were not able to obtain satisfying reconstructions on our datasets using the code provided by the authors.
We report three image quality metrics: mean squared error (MSE; lower is better), structural similarity (SSIM; higher is better)~\cite{Wang04tip}, and the calibrated perceptual loss (LPIPS; lower is better)~\cite{Zhang18cvpr}.

\mypara{Results and Discussion}: The main quantitative results are presented in Table~\ref{tab:image_quality_comparison}, and are supported by qualitative results in Figs.~\ref{fig:comp_event_camera_dataset} and~\ref{fig:comp_bardow}.
Additional results are available in the supplementary material. %
We also encourage the reader to watch the supplementary video, which conveys these results in a better form than still images can.

On all datasets, our reconstruction method outperforms the state-of-the-art by a large margin, with an average 21\% increase in SSIM and a 23\% decrease in LPIPS.
Qualitatively, our method reconstructs small details remarkably well compared to the baselines (see the boxes in the first row of Fig.~\ref{fig:comp_event_camera_dataset}, for example).
Furthermore, our method does not suffer from ``ghosting'' or ``bleeding edges'' artifacts that are present in other methods (particularly visible in the third row of Fig.~\ref{fig:comp_event_camera_dataset}).
These artifacts result from (i) incorrectly estimated contrast thresholds and (ii) the fact that these methods can only estimate the image intensity up to some unknown initial intensity $\Image_0$, whose ghost can remain visible.
We also compare our method to HF, MR, and SOFIE qualitatively using datasets and image reconstructions directly provided by the authors of \cite{Bardow16cvpr}, in Fig.~\ref{fig:comp_bardow}.
Once again, our network generates higher quality reconstructions, with finer details and less noise.
Finally, we show that our network is able to leverage the outstanding properties of events to reconstruct images in low light (Fig.~\ref{fig:qualitative_night}) and during high speed motions (Fig.~\ref{fig:vio_motion_blur_frames}), two scenarios in which conventional cameras fail.

\setlength{\tabcolsep}{0.13ex} %
\global\long\def\heightplot{1.60cm} %
\global\long\def\widthplot{1.60cm} %
\global\long\def\vspacecols{0.2ex} %
\begin{figure}
	\centering
    \begin{tabular}{ccccc}
    \includegraphics[width=\widthplot,height=\heightplot]{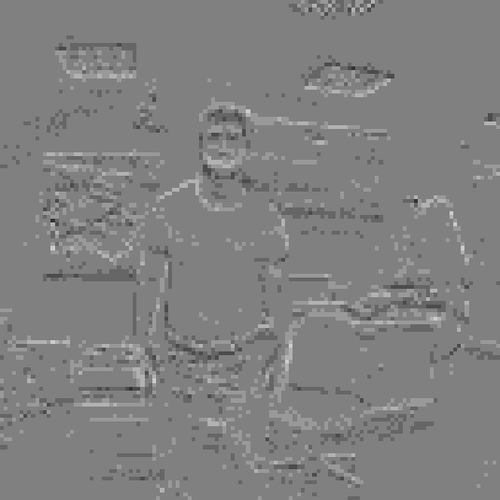}
    & \includegraphics[width=\widthplot,height=\heightplot]{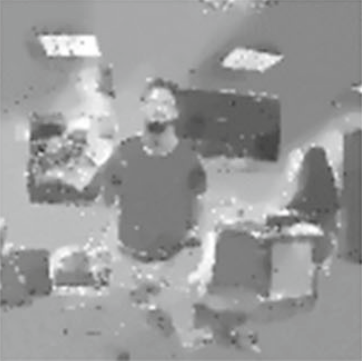}
    & \includegraphics[width=\widthplot,height=\heightplot]{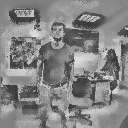}
    & \includegraphics[width=\widthplot,height=\heightplot]{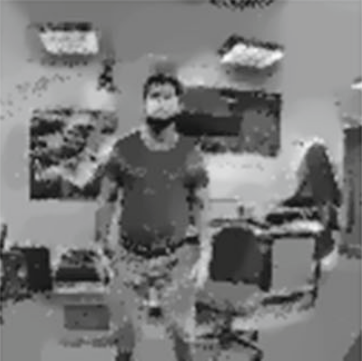}
    & \includegraphics[width=\widthplot,height=\heightplot]{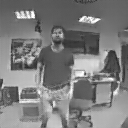}\\
    \includegraphics[width=\widthplot,height=\heightplot]{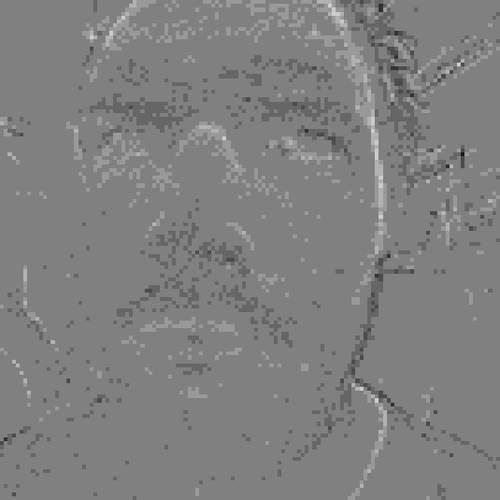}
    & \includegraphics[width=\widthplot,height=\heightplot]{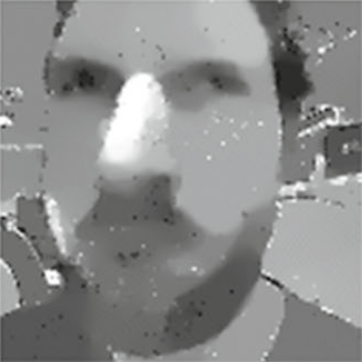}
    & \includegraphics[width=\widthplot,height=\heightplot]{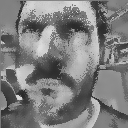}
    & \includegraphics[width=\widthplot,height=\heightplot]{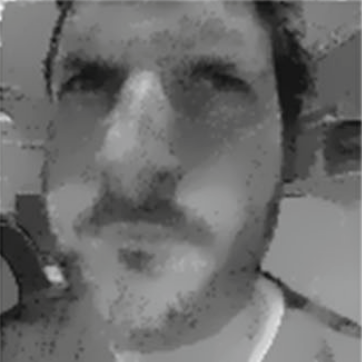}
    & \includegraphics[width=\widthplot,height=\heightplot]{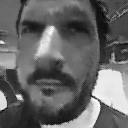}\\
    \footnotesize (a) Events & \footnotesize (b) SOFIE & \footnotesize (c) HF & \footnotesize (d) MR & \footnotesize (e) Ours\\
    \end{tabular}
\vspace{1mm}
\caption{Qualitative comparison on the dataset introduced by \cite{Bardow16cvpr}. Our method produces cleaner and more detailed results.
\vspace{-2ex}
}
\label{fig:comp_bardow}
\end{figure}

\setlength{\tabcolsep}{0.15ex} %
\global\long\def\heightplot{1,852cm} %
\global\long\def\widthplot{2.71cm} %
\begin{figure}
	\centering
    \begin{tabular}{ccc}
    \includegraphics[width=\widthplot,height=\heightplot]{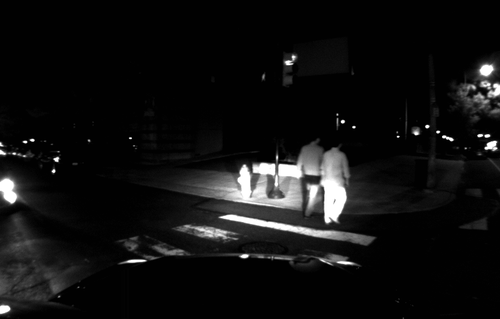}
    & \includegraphics[width=\widthplot,height=\heightplot]{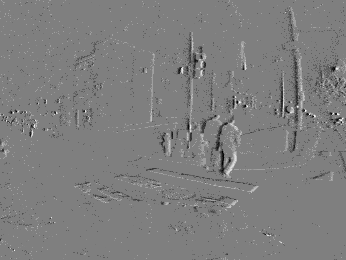}
    & \includegraphics[width=\widthplot,height=\heightplot]{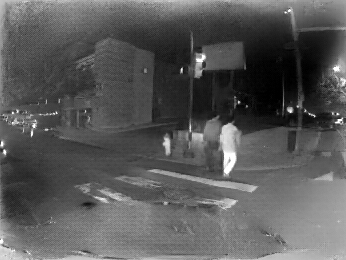}\\
    \footnotesize (a) VI Sensor & \footnotesize (b) Events & \footnotesize (c) Our reconstruction
    \end{tabular}
\vspace{0.5ex}
\caption{A high-dynamic-range reconstruction from an event camera mounted on a car driving at night \cite{Zhu18ral}.
Because of low light, the conventional camera image is severely degraded (a).
In contrast, the events (b) capture the whole dynamic range of the scene, which our method successfully converts into an image (c), recovering details that are lost in the conventional frame. %
}
\label{fig:qualitative_night}
\end{figure}

\begin{table*}
\newcolumntype{Z}{S[table-format=2.2,table-auto-round]}
\centering
\setlength{\tabcolsep}{3mm}
\ra{1.05}
\small
\begin{tabular}{@{}lcZZZcZZZcZZZcr@{}}
  \toprule
  \multirow{2}[3]{*}{Dataset} && \multicolumn{3}{c}{MSE}  &&  \multicolumn{3}{c}{SSIM}  &&\multicolumn{3}{c}{LPIPS} \\
  \cmidrule(l{3mm}r{3mm}){3-5} \cmidrule(l{3mm}r{3mm}){7-9} \cmidrule(l{3mm}r{3mm}){11-13}
  && {HF} & {MR} & {Ours} && {HF} & {MR} & {Ours} && {HF} & {MR} & {Ours}  \\
  \midrule
  dynamic\_6dof && 0.103 & 0.107 & \bfseries 0.080 && 0.394 & 0.439 & \bfseries 0.496 && 0.534 & 0.529 & \bfseries 0.427 \\
boxes\_6dof && 0.087 & 0.074 & \bfseries 0.041 && 0.446 & 0.469 & \bfseries 0.632 && 0.507 & 0.535 & \bfseries 0.360 \\
poster\_6dof && 0.060 & 0.050 & \bfseries 0.037 && 0.521 & 0.550 & \bfseries 0.681 && 0.444 & 0.500 & \bfseries 0.315 \\
shapes\_6dof && 0.111 & 0.142 & \bfseries 0.097 && 0.335 & 0.431 & \bfseries 0.440 && 0.633 & 0.640 & \bfseries 0.531 \\
office\_zigzag && 0.090 & 0.062 & \bfseries 0.048 && 0.361 & 0.425 & \bfseries 0.503 && 0.543 & 0.546 & \bfseries 0.435 \\
slider\_depth && 0.081 & 0.082 & \bfseries 0.058 && 0.476 & 0.511 & \bfseries 0.608 && 0.499 & 0.546 & \bfseries 0.419 \\
calibration && 0.074 & 0.056 & \bfseries 0.039 && 0.405 & 0.411 & \bfseries 0.523 && 0.548 & 0.569 & \bfseries 0.465 \\
\midrule
Mean && 0.086 & 0.082 & \bfseries 0.057 && 0.420 & 0.462 & \bfseries 0.555 && 0.530 & 0.552 & \bfseries 0.422 \\
   \bottomrule
\end{tabular}
\vspace{1mm}
\caption{Comparison to state-of-the-art image reconstruction methods on the Event Camera Dataset \cite{Mueggler17ijrr}. Our approach outperforms prior such methods on all datasets by a large margin, with an average 21\% increase in structural similarity (SSIM) and a 23\% decrease in perceptual distance (LPIPS) compared to the best prior methods, respectively  MR \cite{Munda18ijcv} and HF \cite{Scheerlinck18accv}.}
\label{tab:image_quality_comparison}
\end{table*}

\mypara{Limitations:} Our method introduces some latency due to the fact that we process events in windows as opposed to event-per-event \cite{Scheerlinck18accv}, ranging from \SI{1}{\ms} to \SI{200}{\ms} depending on the event rate and value of $\NumEvents$ \revised{(details in the appendix)}.
Also, in some cases (extreme electronic noise in the events not modelled in simulation), our method can fail to reconstruct correctly some parts of the images, and may propagate the error to the next reconstructions (Fig.~\ref{fig:failure_cases_error_propagation}).

\global\long\def\widthplot{2.03cm} %
\begin{figure}
	\centering
    \begin{tabular}{cccc}
    \includegraphics[width=\widthplot]{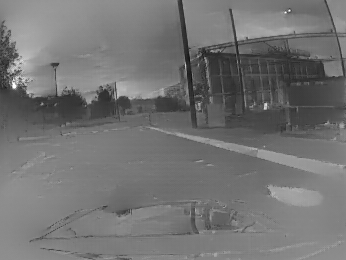}
    & \includegraphics[width=\widthplot]{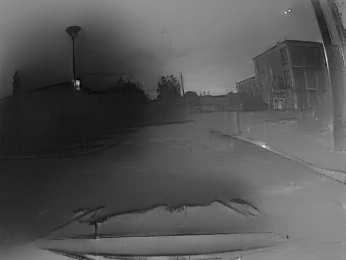}
    & \includegraphics[width=\widthplot]{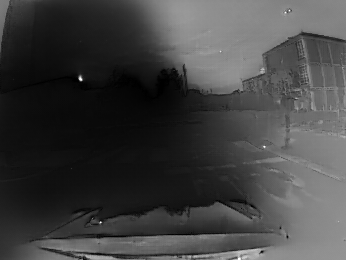}
    & \includegraphics[width=\widthplot]{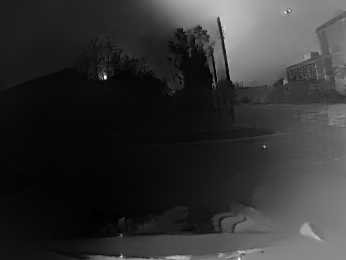}\\
    \footnotesize (a) & \footnotesize (b) & \footnotesize (c) & \footnotesize (d)
    \end{tabular}
\vspace{0.5ex}
\caption{An example failure case on a sequence from the MVSEC dataset \cite{Zhu18ral}.
The sun (top left of (a)) generates noisy events which cause our network to make a local reconstruction error (b), which gets amplified in subsequent reconstructions (c,d).
}
\label{fig:failure_cases_error_propagation}
\end{figure}

%% file: chapters/exp_classification.tex
\section{Downstream Applications}

In this section, we demonstrate the potential of our method as a bridge between conventional computer vision and vision \revised{with event cameras}, for both low-level and high-level tasks.
First, we focus on object classification from events (Section~\ref{sec:exp_classification}) and then turn to camera pose estimation with events and inertial measurements (Section~\ref{sec:exp_vio}).

\subsection{Object Classification}
\label{sec:exp_classification}

Pattern recognition \revised{from event data} is an active research topic.\footnote{A list of related works can be found at: \url{https://github.com/uzh-rpg/event-based_vision_resources}}.
While one line of work focuses on spiking neural architectures (SNNs) to recognize patterns from a stream of events with minimal latency (H-FIRST \cite{Orchard15pami}), conventional machine learning techniques combined with novel event representations such as time surfaces (HOTS~\cite{Lagorce16pami}) have shown the most promising results so far.
Recently, HATS~\cite{Sironi18cvpr} addressed the problem of object classification from a stream of events.
They proposed several modifications to HOTS, and achieved large improvements in classification accuracy, outperforming all prior approaches by a large margin.

We propose an alternative approach to object classification based on a stream of events.
Instead of using hand-crafted event representations, we directly train a classification network on images reconstructed from events.

We compare our approach against several recent methods: HOTS, and the state-of-the-art HATS, using the datasets and metric (classification accuracy) used in the HATS paper.
The N-MNIST (Neuromorphic-MNIST) and N-Caltech101 datasets \cite{Orchard15fns} are event-based versions of the MNIST \cite{Lecun98ieee} and Caltech101 \cite{Li06pami} datasets.
To convert the images to event sequences, an event camera was placed on a motor, and automatically moved while pointing at images from MNIST (respectively Caltech101) that were projected onto a white wall.
The N-CARS dataset \cite{Sironi18cvpr}
proposes a binary classification task: deciding whether a car is visible or not using a \SI{100}{\ms} sequence of events.
Fig.~\ref{fig:preview_classification_datasets} shows a sample event sequence from each of the three datasets.

Our approach follows the same methodology for each dataset.
First, for each event sequence in the training set, we use our network to reconstruct an image from the events (Fig.~\ref{fig:preview_classification_datasets}, bottom row).
Then, we train an off-the-shelf CNN for object classification using the reconstructed images from the training set.
For N-MNIST, we use a simple CNN (details in the supplement) and train it from scratch.
For N-Caltech101 and N-CARS, we use ResNet-18 \cite{He16cvpr}, initialized with weights pretrained on ImageNet \cite{Russakovsky15ijcv}, and fine-tune the network for the dataset at hand.
Once trained, we evaluate each network on the test set (images reconstructed from the events in the test set) and report the classification accuracy.
Furthermore, we perform a transfer learning experiment for the N-MNIST and N-Caltech101 datasets (for which corresponding images are available for every event sequence): we train the CNN on the conventional image datasets, and evaluate the network directly on images reconstructed from events without fine-tuning.

\setlength{\tabcolsep}{0.25cm} %
\global\long\def\heightplot{1.7cm} %
\begin{figure}
	\centering
    \begin{tabular}{ccc}
    \includegraphics[height=\heightplot]{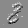}\hspace{1ex}
    & \includegraphics[height=\heightplot]{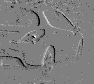}\hspace{1ex}
    & \includegraphics[height=\heightplot]{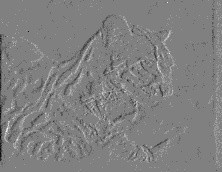}\\
    \includegraphics[height=\heightplot]{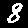}\hspace{1ex}
    & \includegraphics[height=\heightplot]{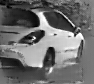}\hspace{1ex}
    & \includegraphics[height=\heightplot]{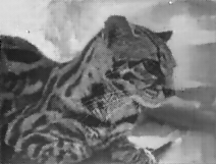}\\
    \footnotesize (a) N-MNIST & \footnotesize (b) N-CARS & \footnotesize (c) N-Caltech101 \\
    \end{tabular}
\vspace{0.5ex}
\caption{Samples from each dataset used in the evaluation of our object classification approach based on events (Section~\ref{sec:exp_classification}). Top: preview of the event sequence. Bottom: our image reconstruction.
}
\label{fig:preview_classification_datasets}
\end{figure}

\begin{table}
\centering
\ra{1.05}
\resizebox{1.0\linewidth}{!}{
	\small
	\begin{tabular}{@{}l@{\hspace{6mm}}*{12}{c@{\hspace{4mm}}}c@{\hspace{8mm}}r@{\hspace{3mm}}r@{}}
		\toprule
		       & N-MNIST & N-CARS & N-Caltech101 \\
		\midrule
		HOTS & 0.808 & 0.624 & 0.210 \\
		HATS/linear SVM & \textbf{0.991} & 0.902 & 0.642 \\
		\revised{HATS/ResNet-18} & \revised{n.a.} & \revised{0.904} & \revised{0.700} \\
		Ours (transfer learning) & 0.807 & n.a. & 0.821 \\
		Ours (fine-tuned) & 0.983 & \textbf{0.910} & \textbf{0.866} \\
		\bottomrule
	\end{tabular}
}
\vspace{1mm}
\caption{Classification accuracy compared to recent approaches, including HATS \cite{Sironi18cvpr}, the state-of-the-art.
}
\label{tab:classification_results}
\vspace{-4mm}
\end{table}

For the baselines, we report directly the accuracy provided in \cite{Sironi18cvpr}. %
\revised{To make the comparison with HATS as fair as possible, we also provide results of classifying HATS features with a ResNet-18 network (instead of the linear SVM used originally).}
The results are presented in Table~\ref{tab:classification_results}, where the datasets are presented in increasing order of difficulty from left to right.
Despite the simplicity of our approach, it outperforms all baselines, and the gap between our method and the state-of-the-art increases as the datasets get more difficult.
While we perform slightly worse than HATS on N-MNIST (98.3\% versus 99.1\%), this can be attributed to the synthetic nature of N-MNIST, for which our approach does not bring substantial advantages compared to a hand-crafted feature representation such as HATS.
Note that, in contrast to HATS, we did not perform hyperparameter tuning.
On N-CARS (binary classification task with natural event data), our method performs better, though the improvement is minor (91\% versus \revised{90.4\%} for HATS).
However, N-CARS is almost saturated in terms of accuracy. %

On N-Caltech101 (the most challenging dataset, requiring classification of natural event data into 101 object classes), our method truly shines, outperforming HATS by a large margin (86.6\% versus \revised{70.0\%}).
This significant gap can be explained by the fact that our approach leverages decades of computer vision research and datasets.
Lifting the event stream into the image domain with our events-to-video approach allows us to use a mature CNN architecture pretrained on existing large, labeled datasets, thus leveraging powerful hierarchical features learned on a large amount of image data -- something that is not possible with event data, for which labeled datasets are scarce.
Finally, and perhaps even more strikingly, we point out that our approach, in the pure transfer learning setup (\ie feeding images reconstructed from events to a network trained on real image data) performs better than all other methods, while not using the event sequences from the training set.
To the best of our knowledge, this is the first time that direct transfer learning between image data and event data has been achieved.

\revised{We also point out that our approach is real-time capable.
On N-Caltech101, end-to-end classification takes less than \SI{10}{\ms} (sequence reconstruction: $\leq\!\SI{8}{\ms}$, object classification: $\leq\!\SI{2}{\ms}$) on an NVIDIA RTX 2080 Ti GPU.
More details about performance can be found in the appendix.}

%% file: chapters/exp_vio.tex
\subsection{Visual-Inertial Odometry}
\label{sec:exp_vio}

\setlength{\tabcolsep}{0.15cm} %
\global\long\def\heightplot{2.625cm} %
\global\long\def\widthplot{3.5cm} %
\begin{figure}[t]
	\centering
    \begin{tabular}{cc}
    \includegraphics[width=\widthplot,height=\heightplot]{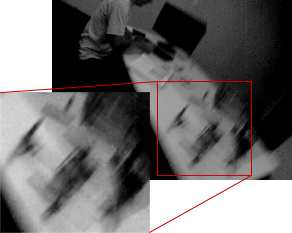}
    &\includegraphics[width=\widthplot,height=\heightplot]{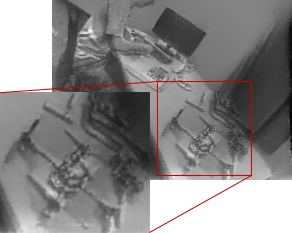}\\
    (a) DAVIS frame & (b) Our reconstruction\\
    \end{tabular}
\vspace{0.5ex}
\caption{Comparison of DAVIS frames and reconstructed frames on a high-speed portion of the `dynamic\_6dof' sequence.
Our reconstructions from events do not suffer from motion blur, which leads to increased pose estimation accuracy (Table~\ref{tab:vio_results}).
}
\vspace{-3mm}
\label{fig:vio_motion_blur_frames}
\end{figure}

\begin{table}[t]
\centering
\ra{1.05}
\resizebox{1.0\linewidth}{!}{
	\small
	\begin{tabular}{@{}l@{\hspace{5mm}}*{12}{c@{\hspace{3mm}}}c@{\hspace{8mm}}r@{\hspace{2mm}}r@{}}
		\toprule
		          & Ours & U.SLAM & U.SLAM & HF & MR & VINS-Mono \\
	    Inputs    & \small E+I & \small E+I & \small E+F+I & \small E+I & \small E+I & \small F+I \\
		\midrule
		shapes\_translation & 0.18 & 0.32 & \textbf{0.17} & failed & 2.00 & 0.93 \\
		poster\_translation & \textbf{0.05} & 0.09 & 0.06 & 0.49 & 0.15 & failed\\
		boxes\_translation & \textbf{0.15} & 0.81 & 0.26 & 0.70 & 0.45 & 0.22\\
		dynamic\_translation & \textbf{0.08} & 0.23 & 0.09 & 0.58 & 0.17 & 0.13\\
		shapes\_6dof & 1.09 & 0.09 & \textbf{0.06} & failed & 3.00 & 1.99\\
		poster\_6dof & \textbf{0.12} & 0.20 & 0.22 & 0.45 & 0.17 & 1.99\\
		boxes\_6dof & 0.62 & 0.41 & \textbf{0.34} & 1.71 & 1.17 & 0.94\\
		dynamic\_6dof & 0.15 & 0.27 & \textbf{0.11} & failed & 0.55 & 0.76\\
		hdr\_boxes & 0.34 & 0.44 & 0.37 & 0.64 & 0.66 & \textbf{0.32}\\
		\midrule
		Mean & 0.31 & 0.32 & \textbf{0.19} & 0.76 & 0.92 & 0.91\\
		Median & \textbf{0.15} & 0.27 & 0.17 & 0.61 & 0.55 & 0.84\\
		\bottomrule
	\end{tabular}
}
\vspace{1mm}
\caption{Mean translation error (in meters) on the sequences from~\cite{Mueggler17ijrr}.
Our method outperforms all other methods that use events and IMU, including UltimateSLAM~(E+I).
Surprisingly, it even performs on par with UltimateSLAM (E+F+I), while not using additional frames. Methods for which the mean translation error exceeds $\SI{5}{\meter}$ are marked as ``failed''. %
}
\label{tab:vio_results}
\vspace{-3mm}
\end{table}

The task of Visual-inertial odometry (VIO) is to recover the 6-degrees-of-freedom (6-DOF) pose of a camera from a set of visual measurements (images or events) and inertial measurements from an inertial measurement unit (IMU) that is rigidly attached to the camera.
Because of its importance in augmented/virtual reality and mobile robotics, VIO has been extensively studied in the last decade and is relatively mature today \cite{Mourikis07icra,Leutenegger15ijrr,Blosch15iros,Forster17troOnmanifold,Qin17arxiv}.
Yet systems based on conventional cameras fail in challenging conditions such as high-speed motions or high-dynamic-range environments. This has recently motivated the development of VIO systems \revised{with event data} (EVIO) \cite{Zhu17cvpr,Rebecq17bmvc,Rosinol18ral}.

The state-of-the-art EVIO system, UltimateSLAM~\cite{Rosinol18ral}, operates by independently tracking visual features from pseudo-images reconstructed from events and optional images from a conventional camera, and fusing the tracks with inertial measurements using an existing optimization backend \cite{Leutenegger15ijrr}.
Here, we go one step further and directly apply an off-the-shelf VIO system (specifically, VINS-Mono \cite{Qin17arxiv}, which is state-of-the-art \cite{Delmerico18icra}) to videos reconstructed from events using either our approach, MR, or HF, and evaluate against UltimateSLAM.
As is standard \cite{Zhu17cvpr,Rebecq17bmvc,Rosinol18ral}, we use sequences from the Event Camera Dataset \cite{Mueggler17ijrr}, which contain events, frames, and IMU measurements from a DAVIS240C \cite{Brandli14ssc} sensor.
Each sequence is 60 seconds long, and contains data from a hand-held event camera undergoing a variety of motions in several environments.
All sequences feature extremely fast motions (angular velocity up to \SI{880}{\degree/\second} and linear velocity up to \SI{3.5}{\meter/\second}), which leads to severe motion blur on the frames (Fig.~\ref{fig:vio_motion_blur_frames}).
We compare our approach against the two operating modes of UltimateSLAM: UltimateSLAM (E+I) which uses only events and IMU, and UltimateSLAM (E+F+I) that uses the events, the IMU, and additional frames. %
We run a publicly available VIO evaluation toolbox \cite{Zhang18iros} on raw trajectories provided by the authors of UltimateSLAM, which ensures that the trajectories estimated by all methods are evaluated in the exact same manner.
For completeness, we also report results from running VINS-Mono directly on the frames from the DAVIS sensor.

Table \ref{tab:vio_results} presents the mean translation error of each method, for all datasets (additional results are presented in the supplement).
First, we note that our method performs better than UltimateSLAM (E+I) on all sequences, with the exception of the `shapes\_6dof' sequence.
This sequence features a few synthetic shapes with very few features ($\leq\!10$), which cause VINS-Mono to not properly initialize, leading to high error (note that this is a problem with VINS-Mono and not our image reconstructions).
Overall, the median error of our method is \SI{0.15}{\meter}, which is almost twice smaller than UltimateSLAM (E+I) (\SI{0.27}{\meter}), which uses the exact same data. %
Indeed, while UltimateSLAM (E+I) uses coarse pseudo-images created from a single, small window of events, our network is able to reconstruct images with finer details, and higher temporal consistency -- both of which lead to better feature tracks, and thus better pose estimates.
Even more strikingly, our approach performs on par with UltimateSLAM (E+F+I), while the latter requires additional frames which we do not need.
The median error of both methods is comparable (\SI{0.15}{\meter} for ours versus \SI{0.17}{\meter} for UltimateSLAM (E+F+I)).%

Finally, we point out that running the same VIO (VINS-Mono) on competing image reconstructions (MR and HF) leads to significantly larger tracking errors (e.g. median error three times as large for MR), which further highlights the superiority of our image reconstructions for downstream vision applications.
We acknowledge that our approach is not as fast as UltimateSLAM. %
Since the main difference between both approaches is how they convert events into ``image-like'' representations, a rough estimate of the performance gap can be obtained by comparing the time it takes for each method to synthesize a new image: UltimateSLAM takes about $\SI{1}{\ms}$ on a CPU, \revised{in comparison to $\leq\!\SI{4}{\ms}$ on a high-end GPU for our method.}
Nevertheless, our events-to-video network allows harnessing the outstanding properties of events for VIO, reaching even higher accuracy than state-of-the-art EVIO designed specifically for event data.

%% file: chapters/conclusion.tex
\vspace{-1ex}
\section{Conclusion}

We presented a novel events-to-video reconstruction framework based on a recurrent convolutional network trained on simulated event data.
In addition to outperforming state-of-the-art reconstruction methods on real event data by a large margin ($>\!20\%$ improvement), we showed the applicability of our method as a bridge between conventional cameras and event cameras on two vision applications, namely object classification from events and visual-inertial odometry.
For each of these tasks, we applied an off-the-shelf computer vision algorithm to videos reconstructed from events by our network, and showed that the result outperforms state-of-the-art algorithms tailored for event data in each case. This validates that our approach allows to readily apply decades of computer vision research to event cameras: mature algorithms, modern deep architectures, and weights pretrained from large image datasets. %

%% file: chapters/supplement.tex
\section{\revised{Architecture Details}}

\begin{figure*}[t]
	\centering
    \includegraphics[width=\linewidth]{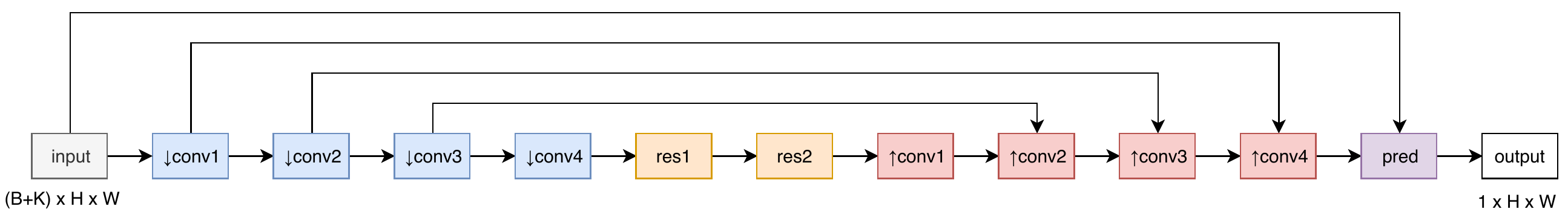}
\vspace{0.1ex}
\caption{\revised{Detailed architecture of our reconstruction network. It is a UNet architecture \cite{Ronneberger15icmicci}, composed of multiple convolutional layers. Four encoder layers (blue) are followed by two residual blocks (yellow), four decoder layers (red), and a final image prediction layer. In addition, symmetric skip connections are used. More details about each layer are provided in the text.}}
\label{fig:architecture_details}
\end{figure*}

\revised{Our reconstruction network is presented in detail in Fig.~\ref{fig:architecture_details}.
It is essentially a UNet architecture \cite{Ronneberger15icmicci}, composed of multiple convolutional layers.
Four encoder layers (blue) are followed by two residual blocks (yellow), four decoder layers (red), and a final image prediction layer.
In addition, symmetric skip connections are used.
The encoders are strided convolutional layers (stride of 2), with a kernel size of 5.
The number of output channels of the first encoder layer ($\downarrow$conv1) is 64, and is doubled for every subsequent encoder layer, \ie the sequence of output channels is $(64,128,256,512)$.
Both residual blocks have 512 hidden layers, and a kernel size of 3.
Batch normalization is used within the residual blocks (applied before the activation).
The decoders are transposed convolution layers, with a kernel size of 5.
The number of output channels of the decoders starts at 256 ($\uparrow$conv1), and is divided by two for every subsequent decoder, \ie the sequence of output channels is $(256, 128, 64, 32)$).
The kernel size for the residual blocks is 3.
ReLU is used as activation everywhere, except for the last image prediction layer, where a sigmoid activation is used instead.
The skip connections are based on concatenation.}

\section{Initialization Phase}
\label{sec:initialization_phase}

By analyzing the initialization phase (\ie when only few events have been triggered yet) in detail we gain interesting insight into how our network operates. We see significantly different behaviour when compared to prior approaches that are based on direct event integration.
Fig.~\ref{fig:analysis_burnin_period_supp} compares image reconstructions from our approach, HF, and MR during the initialization phase. We specifically examine the interval from $\SI{0}{\second}$ to $\SI{0.5}{\second}$ after the event cameras has been started.

HF and MR, which rely on event integration, can only recover the intensity up to the initial (unknown) image $\Image_0$ (\ie they can only recover $\Reconstructed \approx \Image-\Image_0$), which results in an ``edge'' image which does not capture the appearance of the scene correctly.
In contrast, our method successfully leverages deep priors to reconstruct the scene despite the low number of events that is available.

\setlength{\tabcolsep}{0.3ex} %
\global\long\def\heightplot{2,55cm} %
\global\long\def\widthplot{3.4cm} %
\global\long\def\vspacecols{0.3ex} %
\begin{figure*}
	\centering
    \begin{tabular}{cccc}
    \includegraphics[width=\widthplot,height=\heightplot]{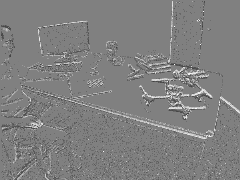}
    & \includegraphics[width=\widthplot,height=\heightplot]{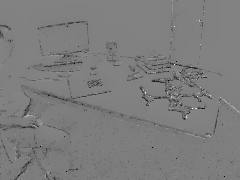}
    & \includegraphics[width=\widthplot,height=\heightplot]{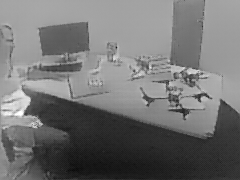}
    & \includegraphics[width=\widthplot,height=\heightplot]{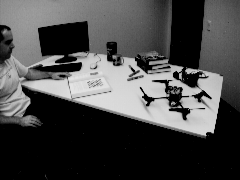}\\[\vspacecols]
    \includegraphics[width=\widthplot,height=\heightplot]{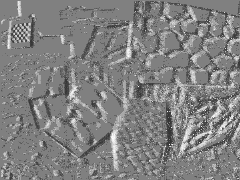}
    & \includegraphics[width=\widthplot,height=\heightplot]{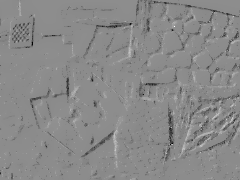}
    & \includegraphics[width=\widthplot,height=\heightplot]{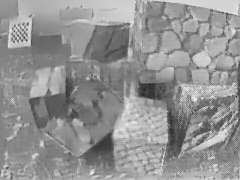}
    & \includegraphics[width=\widthplot,height=\heightplot]{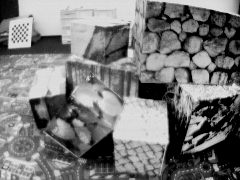}\\[\vspacecols]
    \includegraphics[width=\widthplot,height=\heightplot]{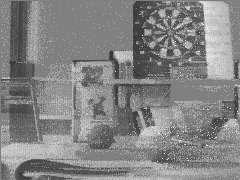}
    & \includegraphics[width=\widthplot,height=\heightplot]{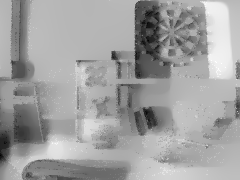}
    & \includegraphics[width=\widthplot,height=\heightplot]{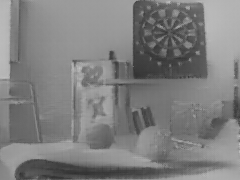}
    & \includegraphics[width=\widthplot,height=\heightplot]{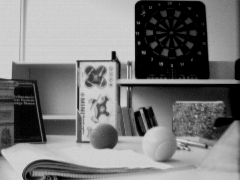}\\[\vspacecols]
    (a) HF & (b) MR & (c) Ours & (d) Ground truth\\
    \end{tabular}
\vspace{0.5ex}
\caption{Analysis of the initialization phase (reconstruction from few events). This figure shows image reconstructions from each method, 0.5 seconds after the sensor was started. HF \cite{Scheerlinck18accv} and MR \cite{Munda18ijcv}, which are based on event integration, cannot recover the intensity correctly, but only the intensity difference , resulting in ``edge '' images (first and second row), or severe ``ghosting'' effects (third row, where the trace of the dartboard is clearly visible).
In contrast, our network successfully reconstructs most of the scene accurately, even with a low number of events.
}
\label{fig:analysis_burnin_period_supp}
\end{figure*}

By contrast, our method is able to leverage deep priors on what a scene might look like (learned, first, from the large amount of simulated event data, and second, embedded in the perceptual loss used (LPIPS)) to reconstruct the scene reasonably well even with a low number of events.

\section{\revised{Why Use Synthetic Training Data?}}

\begin{figure}[h!]
\begin{center}
   \includegraphics[width=0.49\linewidth]{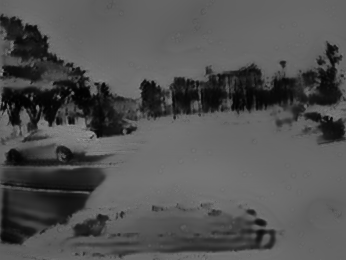}
   \includegraphics[width=0.49\linewidth]{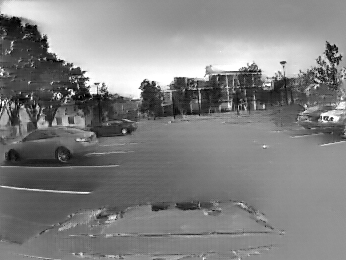}
\end{center}
   \vspace{-0.4cm}
   \caption{\revised{Reconstruction from (i) a network trained only on real event data from the DAVIS240C sensor (left), and (ii) a network trained only on simulated event data (right). This sequence is from the MVSEC dataset, and was recorded with a mDAVIS346 sensor.}}
\label{fig:training_real_versus_sim}
\end{figure}

\revised{Here, we expand on the reasons that motivated us to train our reconstruction network using synthetic event data.
First, simulation allows to capture a large variety of scenes and motions at very little cost.
Second, a conventional camera (even a high quality one) would provide poor ground truth in high-speed conditions (motion blur) and HDR scenes, which are the conditions in which event sensors excel; by contrast, synthetic data does not suffer from these issues.
Last but not least, simulation allows to randomize the contrast thresholds of the event sensor, which increases the ability of the network to generalize to different sensor configurations (contrast sensitivity).
To illustrate this last point, we show in Fig.~\ref{fig:training_real_versus_sim} (left) what happens when training the network on real event data from an event  camera (specifically, the sequences from the Event Camera Dataset \cite{Mueggler17ijrr} already presented in the main paper, which were recorded with a DAVIS240C sensor), and evaluating the trained network on data coming from a different event sensor (specifically, the `outdoors\_day1` sequence from the MVSEC dataset \cite{Zhu18ral}, which was recorded with a mDAVIS346 sensor): the reconstruction suffers from many artefacts.
This can be explained by the fact that the events from the mDAVIS346 sensor have statistics that are quite different from the training events (DAVIS240C): the set of contrast thresholds are likely quite different between both sensors, and the illumination conditions are also different (outdoor lighting for the MVSEC dataset versus indoor lighting for the training event data).
By contrast, the network trained on simulated event data (Fig.~\ref{fig:training_real_versus_sim}, right) generalizes well to the event data from the mDAVIS346, producing a visually pleasing image reconstruction.}

\section{\revised{Latency and Performance Considerations}}

\subsection{Latency}
\label{sec:latency}

\begin{figure*}
	\centering
	\includegraphics[width=0.7\linewidth]{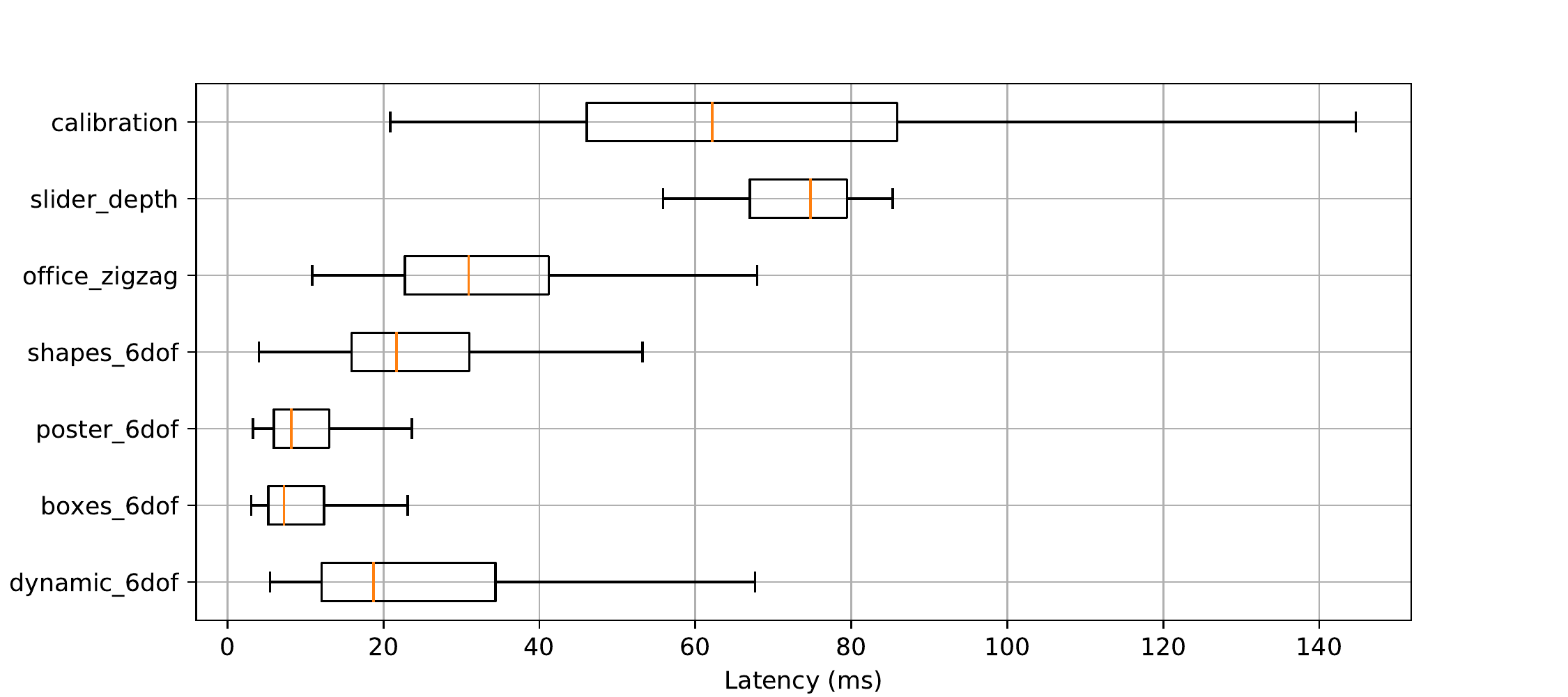}
	\caption{Distribution of latency for each dataset used in our quantitative evaluation.}
\label{fig:boxplot_latency}
\end{figure*}

Our method operates on windows containing a fixed number of events, which incurs some latency compared to approaches that operate in an event-by-event fashion.
The amount of latency depends on the duration of the event windows.
Therefore, it varies through time, depending on the event rate.
Since we process events in windows with a constant number of events, our method is data-driven, which means that the latency diminishes when the event rate increases, \ie when something of interest occurs in the scene.
Fig.~\ref{fig:boxplot_latency} shows the distribution of the latency (in other words, the durations of the event windows) for the sequences used in our qualitative evaluation (Section~4 in the main paper): the latency is in the range of \SI{3}{\ms} to \SI{300}{\ms}, with a median value of \SI{26}{\ms}.
We point out that while in this paper we chose to process windows with a fixed number of events, our methodology could also be applied to windows of a constant duration (which would require, however, to retrain the network to operate with such fixed-length windows), yielding a constant, predictable latency, but at the cost of losing the adaptive output framerate.

\subsection{\revised{Performance Analysis}}
\label{sec:performance_comparison}

\revised{
In this section, we analyze the performance of our method, and compare it against HF \cite{Scheerlinck18accv} and MR \cite{Munda18ijcv}.
Due to fundamental differences in the way each of these methods process event data, it is difficult to provide a direct and fair performance comparison between these three methods.
HF processes the event stream in an event-by-event fashion, providing (in theory) a new image reconstruction with every incoming event.
However, the raw image reconstructions from HF need to be filtered (for example, using a bilateral filter) to obtain results with reasonable quality.
While MR can in principle also operate in an event-by-event fashion, its best quality results are obtained when it processes small batches of events (in our experiments, we used $N_{MR}=1@000$ events per batch), thus limiting the output framerate.
Our method, by contrast, processes the event stream in large batches of events (we used $N=25@000$ events per batch), thus also limiting the output framerate (\ie increasing the latency, as analyzed in Fig.~\ref{fig:boxplot_latency}).
In Table~\ref{tab:performance_comparison}, we report the mean event processing rate (\ie the total time it takes to process a dataset divided by the number of events in the dataset) for each method.
As a complementary performance measure, we also report the mean ``frame synthesis time'', which we define as follows:
\begin{itemize}
  \item for our method, it is the time it takes to process $N$ events.
  \item For MR, it is the time it takes to process $N_{MR}$ events.
  \item For HF, it is in theory the time it takes to process a single event (since every new event triggers a new reconstruction), plus the image filtering time. However, the time to process a single event is negligible compared to the filtering time (multiple orders of magnitude less), hence we report only the filtering time. As described in the main text, we used a bilateral filter with filter size $d=5$ and $\sigma=25$).
\end{itemize}
We ran our method and MR on an NVIDIA GeForce RTX 2080 Ti GPU, and HF on an Intel Core i9-9900K @ 3.60 GHz CPU.
}

\begin{table}[h!]
\centering
\ra{1.05}
\resizebox{1.0\linewidth}{!}{
	\small
	\begin{tabular}{@{}l@{\hspace{6mm}}*{12}{l@{\hspace{4mm}}}l@{\hspace{8mm}}l@{\hspace{3mm}}l@{}}
		\toprule
		       & Event rate (Mev/s) & Frame synthesis time (ms) \\
		\midrule
		HF & 14.30 & 0.75 (filter) \\
		MR & 1.19 & 0.84 \\
	    Ours & 7.94 &  3.15 \\
		\bottomrule
	\end{tabular}
}
\vspace{1mm}
\caption{\revised{Performance comparison between our method, HF, and MR.}}
\label{tab:performance_comparison}
\vspace{-2ex}
\end{table}

\paragraph{Discussion}
\revised{We point out that our method synthesizes fewer images per second than MR and HF, hence the numbers in Table~\ref{tab:performance_comparison} should be interpreted carefully.
That being said, our method is fairly competitive in terms of performance, and, importantly, can easily run in real-time, while providing state of the art reconstructions in terms of quality.}

\section{Additional Results}
\label{sec:supplementary_materials}

\subsection{Video}

We strongly encourage the reader to view the supplemental video, which contains:

\begin{itemize}
  \item Video reconstructions from our method on various event datasets, with a visual comparison to several state of the art methods.
  \vspace{-1ex}
  \item Video of the VINS-Mono visual-inertial odometry algorithm \cite{Qin17arxiv} running on a video reconstruction from events.
  \vspace{-1ex}  
  \item Qualitative results on two additional downstream applications that were not presented in the main paper: object detection (based on YOLOv3 \cite{redmon2018arxiv}), and monocular depth prediction (based on MegaDepth~\cite{Li18cvpr}). We point out that neither of these tasks have ever been shown with event data before this work.
  \end{itemize}
  
\subsection{Results on Synthetic Event Data}
We show a quantitative comparison of the reconstruction quality of our method as well as  MR and HF on synthetic event sequences in Table~\ref{tab:image_quality_comparison_synthetic_supp}. We present qualitative reconstruction results on this dataset in Fig.~\ref{fig:comp_synthetic_supp}.
All methods perform better on synthetic data than real data. This is expected because simulated events are free of noise.
Nonetheless, the performance gap between our method and the state of the art is preserved, and even slightly increases (24\% improvement in SSIM, 56\% decrease in LPIPS).
We note that perfect reconstruction, even on noise-free event streams is not possible, since image reconstruction from events
is only possibly up the the quantization limit imposed by the contrast threshold of the event camera.

\begin{table*}
\newcolumntype{Z}{S[table-format=2.2,table-auto-round]}
\centering
\setlength{\tabcolsep}{3mm}
\ra{1.05}
\small
\begin{tabular}{@{}lcZZZcZZZcZZZcr@{}}
  \toprule
  \multirow{2}[3]{*}{Dataset} && \multicolumn{3}{c}{MSE}  &&  \multicolumn{3}{c}{SSIM}  &&\multicolumn{3}{c}{LPIPS} \\
  \cmidrule(l{3mm}r{3mm}){3-5} \cmidrule(l{3mm}r{3mm}){7-9} \cmidrule(l{3mm}r{3mm}){11-13}
  && {HF} & {MR} & {Ours} && {HF} & {MR} & {Ours} && {HF} & {MR} & {Ours}  \\
  \midrule
synthetic\_0 && 0.082 & 0.055 & \bfseries 0.015 && 0.536 & 0.608 & \bfseries 0.827 && 0.492 & 0.466 & \bfseries 0.260 \\
synthetic\_1 && 0.151 & 0.135 & \bfseries 0.062 && 0.384 & 0.448 & \bfseries 0.598 && 0.536 & 0.559 & \bfseries 0.366 \\
synthetic\_2 && 0.069 & 0.078 & \bfseries 0.022 && 0.603 & 0.676 & \bfseries 0.816 && 0.423 & 0.415 & \bfseries 0.257 \\
synthetic\_3 && 0.067 & 0.050 & \bfseries 0.038 && 0.574 & 0.661 & \bfseries 0.749 && 0.447 & 0.426 & \bfseries 0.327 \\
synthetic\_4 && 0.078 & 0.058 & \bfseries 0.021 && 0.617 & 0.667 & \bfseries 0.854 && 0.406 & 0.418 & \bfseries 0.250 \\
synthetic\_5 && 0.077 & 0.077 & \bfseries 0.024 && 0.501 & 0.606 & \bfseries 0.743 && 0.525 & 0.538 & \bfseries 0.357 \\
synthetic\_6 && 0.068 & 0.039 & \bfseries 0.025 && 0.557 & 0.645 & \bfseries 0.765 && 0.439 & 0.476 & \bfseries 0.298 \\
\bottomrule
Mean && 0.085 & 0.070 & \bfseries 0.030 && 0.539 & 0.616 & \bfseries 0.765 && 0.467 & 0.471 & \bfseries 0.302 \\
   \bottomrule
\end{tabular}
\vspace{1mm}
\caption{Comparison of image quality with respect to state of the art on synthetic event sequences.}
\label{tab:image_quality_comparison_synthetic_supp}
\end{table*}

\setlength{\tabcolsep}{0.3ex} %
\global\long\def\heightplot{2.55cm} %
\global\long\def\widthplot{3.4cm} %
\global\long\def\vspacecols{0.3ex} %
\begin{figure*}
	\centering
    \begin{tabular}{ccccc}
    \includegraphics[width=\widthplot,height=\heightplot]{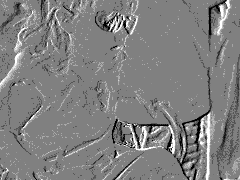}
    & \includegraphics[width=\widthplot,height=\heightplot]{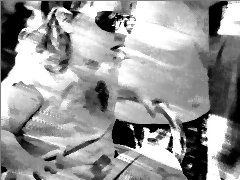}
    & \includegraphics[width=\widthplot,height=\heightplot]{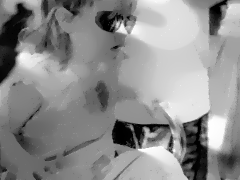}
    & \includegraphics[width=\widthplot,height=\heightplot]{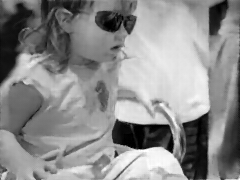}
    & \includegraphics[width=\widthplot,height=\heightplot]{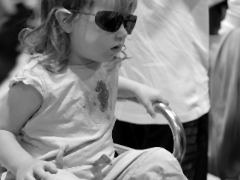}\\[\vspacecols]
    \includegraphics[width=\widthplot,height=\heightplot]{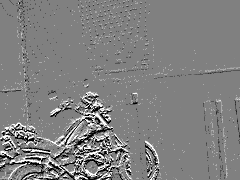}
    & \includegraphics[width=\widthplot,height=\heightplot]{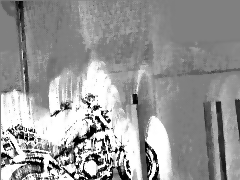}
    & \includegraphics[width=\widthplot,height=\heightplot]{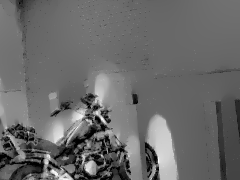}
    & \includegraphics[width=\widthplot,height=\heightplot]{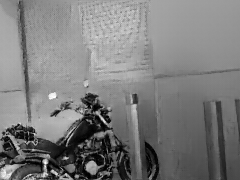}
    & \includegraphics[width=\widthplot,height=\heightplot]{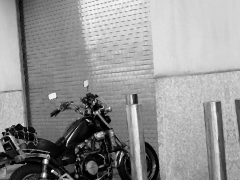}\\[\vspacecols]
    \includegraphics[width=\widthplot,height=\heightplot]{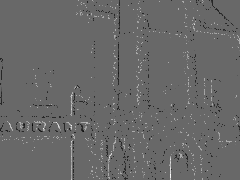}
    & \includegraphics[width=\widthplot,height=\heightplot]{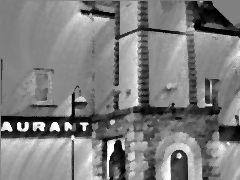}
    & \includegraphics[width=\widthplot,height=\heightplot]{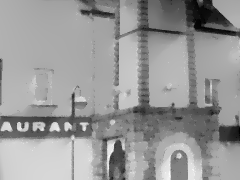}
    & \includegraphics[width=\widthplot,height=\heightplot]{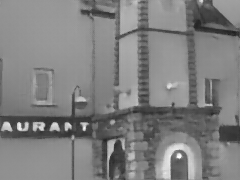}
    & \includegraphics[width=\widthplot,height=\heightplot]{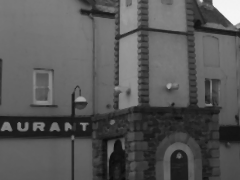}\\[\vspacecols]
    \includegraphics[width=\widthplot,height=\heightplot]{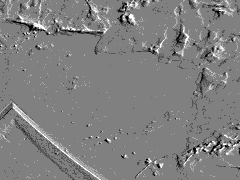}
    & \includegraphics[width=\widthplot,height=\heightplot]{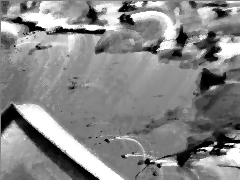}
    & \includegraphics[width=\widthplot,height=\heightplot]{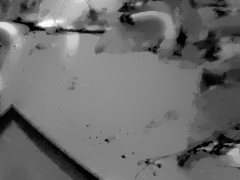}
    & \includegraphics[width=\widthplot,height=\heightplot]{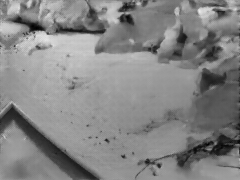}
    & \includegraphics[width=\widthplot,height=\heightplot]{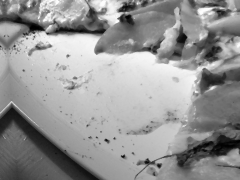}\\[\vspacecols]
    \includegraphics[width=\widthplot,height=\heightplot]{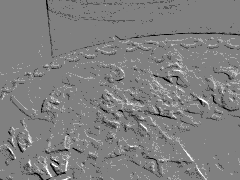}
    & \includegraphics[width=\widthplot,height=\heightplot]{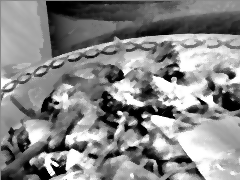}
    & \includegraphics[width=\widthplot,height=\heightplot]{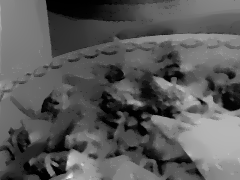}
    & \includegraphics[width=\widthplot,height=\heightplot]{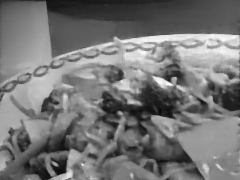}
    & \includegraphics[width=\widthplot,height=\heightplot]{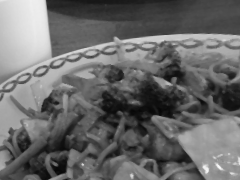}\\[\vspacecols]
    \includegraphics[width=\widthplot,height=\heightplot]{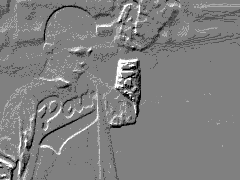}
    & \includegraphics[width=\widthplot,height=\heightplot]{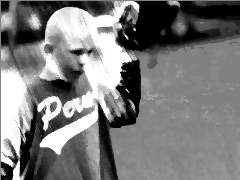}
    & \includegraphics[width=\widthplot,height=\heightplot]{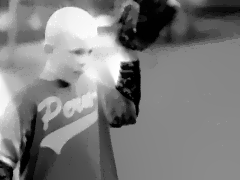}
    & \includegraphics[width=\widthplot,height=\heightplot]{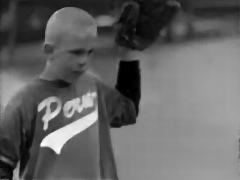}
    & \includegraphics[width=\widthplot,height=\heightplot]{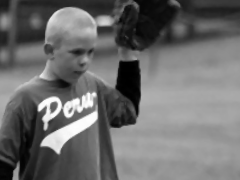}\\[\vspacecols]
    \includegraphics[width=\widthplot,height=\heightplot]{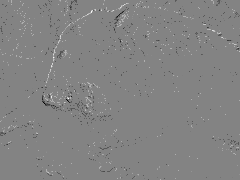}
    & \includegraphics[width=\widthplot,height=\heightplot]{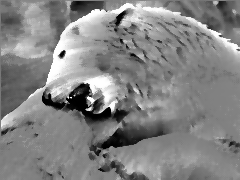}
    & \includegraphics[width=\widthplot,height=\heightplot]{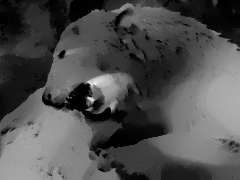}
    & \includegraphics[width=\widthplot,height=\heightplot]{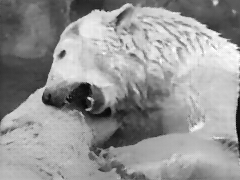}
    & \includegraphics[width=\widthplot,height=\heightplot]{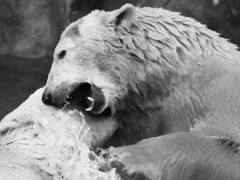}\\[\vspacecols]
    (a) Events & (b) HF & (c) MR & (d) Ours & (e) Ground truth\\
    \end{tabular}
\vspace{1ex}
\caption{Qualitative comparison of our reconstruction method with HF \cite{Scheerlinck18accv} and MR \cite{Munda18ijcv} on synthetic sequences from the validation set. Note our method is able to reconstruct fine details such as the bear's fur (last row), which competing methods are not able to preserve.
}
\label{fig:comp_synthetic_supp}
\vspace{6ex}
\end{figure*}

\subsection{Additional Qualitative Results on Real Data}

Fig.~\ref{fig:comp_event_camera_dataset_supp} shows qualitative results on sequences from the Event Camera Dataset \cite{Mueggler17ijrr} (which we used for our quantitative evaluation). Fig~\ref{fig:comp_bardow_supp} shows qualitative results on the sequences introduced by Bardow \etal \cite{Bardow16cvpr}. Figs.~\ref{fig:qualitative_hdr_supp} and~\ref{fig:qualitative_night_supp} present HDR reconstruction results on sequences from the MVSEC dataset \cite{Zhu18ral}.
Further results are  shown in the supplementary video which conveys these results in a better form than still images.

\setlength{\tabcolsep}{0.3ex} %
\global\long\def\heightplot{2.125cm} %
\global\long\def\widthplot{2.8cm} %
\global\long\def\vspacecols{0.3ex} %
\begin{figure*}
	\centering
    \begin{tabular}{cccccc}
    \includegraphics[width=\widthplot,height=\heightplot]{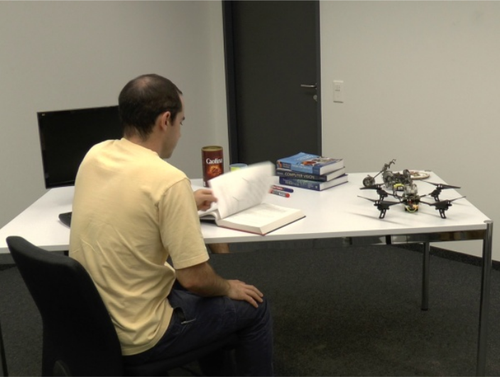}
    & \includegraphics[width=\widthplot,height=\heightplot]{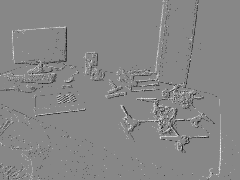}
    & \includegraphics[width=\widthplot,height=\heightplot]{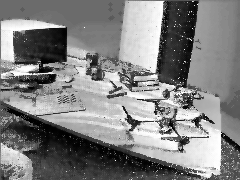}
    & \includegraphics[width=\widthplot,height=\heightplot]{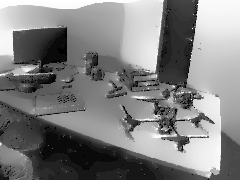}
    & \includegraphics[width=\widthplot,height=\heightplot]{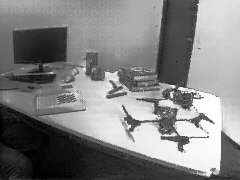}
    & \includegraphics[width=\widthplot,height=\heightplot]{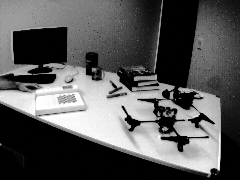}\\[\vspacecols]
    \includegraphics[width=\widthplot,height=\heightplot]{images/comp_event_camera_dataset/preview_boxes.png}
    & \includegraphics[width=\widthplot,height=\heightplot]{images/comp_event_camera_dataset/boxes_6dof/130_events.png}
    & \includegraphics[width=\widthplot,height=\heightplot]{images/comp_event_camera_dataset/boxes_6dof/processed/130_CF.png}
    & \includegraphics[width=\widthplot,height=\heightplot]{images/comp_event_camera_dataset/boxes_6dof/processed/130_MR.png}
    & \includegraphics[width=\widthplot,height=\heightplot]{images/comp_event_camera_dataset/boxes_6dof/processed/130_Ours.png}
    & \includegraphics[width=\widthplot,height=\heightplot]{images/comp_event_camera_dataset/boxes_6dof/processed/130_groundtruth.png}\\[\vspacecols]
    \includegraphics[width=\widthplot,height=\heightplot]{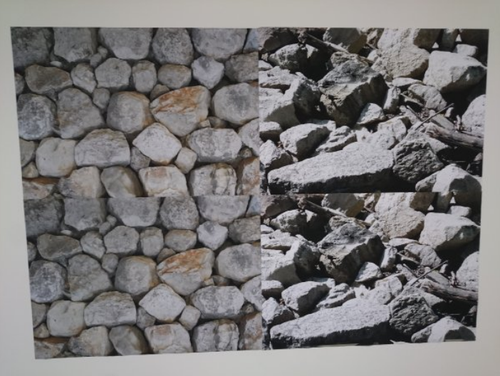}
    & \includegraphics[width=\widthplot,height=\heightplot]{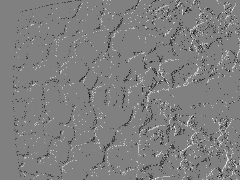}
    & \includegraphics[width=\widthplot,height=\heightplot]{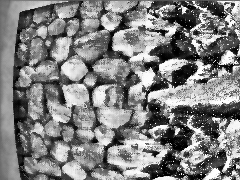}
    & \includegraphics[width=\widthplot,height=\heightplot]{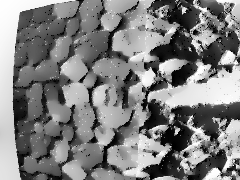}
    & \includegraphics[width=\widthplot,height=\heightplot]{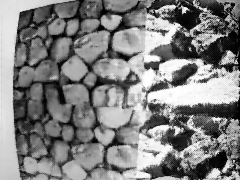}
    & \includegraphics[width=\widthplot,height=\heightplot]{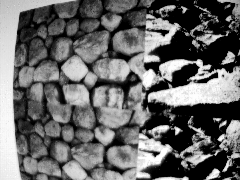}\\[\vspacecols]
    \includegraphics[width=\widthplot,height=\heightplot]{images/comp_event_camera_dataset/preview_shapes.png}
    & \includegraphics[width=\widthplot,height=\heightplot]{images/comp_event_camera_dataset/shapes_6dof/136_events.png}
    & \includegraphics[width=\widthplot,height=\heightplot]{images/comp_event_camera_dataset/shapes_6dof/raw/136_CF.png}
    & \includegraphics[width=\widthplot,height=\heightplot]{images/comp_event_camera_dataset/shapes_6dof/raw/136_MR.png}
    & \includegraphics[width=\widthplot,height=\heightplot]{images/comp_event_camera_dataset/shapes_6dof/raw/136_Ours.png}
    & \includegraphics[width=\widthplot,height=\heightplot]{images/comp_event_camera_dataset/shapes_6dof/raw/136_groundtruth.png}\\[\vspacecols]
    \includegraphics[width=\widthplot,height=\heightplot]{images/comp_event_camera_dataset/preview_office.png}
    & \includegraphics[width=\widthplot,height=\heightplot]{images/comp_event_camera_dataset/office_zigzag/104_events.png}
    & \includegraphics[width=\widthplot,height=\heightplot]{images/comp_event_camera_dataset/office_zigzag/processed/104_CF.png}
    & \includegraphics[width=\widthplot,height=\heightplot]{images/comp_event_camera_dataset/office_zigzag/processed/104_MR.png}
    & \includegraphics[width=\widthplot,height=\heightplot]{images/comp_event_camera_dataset/office_zigzag/processed/104_Ours.png}
    & \includegraphics[width=\widthplot,height=\heightplot]{images/comp_event_camera_dataset/office_zigzag/processed/104_groundtruth.png}\\[\vspacecols]
    \includegraphics[width=\widthplot,height=\heightplot]{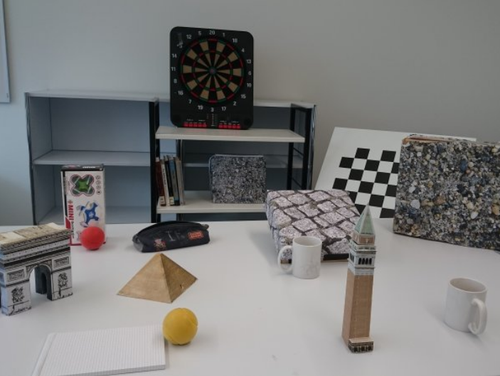}
    & \includegraphics[width=\widthplot,height=\heightplot]{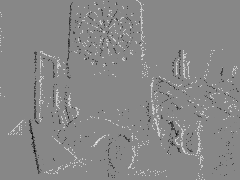}
    & \includegraphics[width=\widthplot,height=\heightplot]{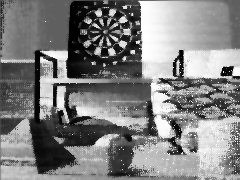}
    & \includegraphics[width=\widthplot,height=\heightplot]{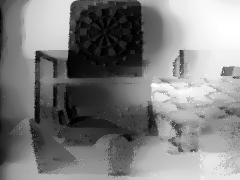}
    & \includegraphics[width=\widthplot,height=\heightplot]{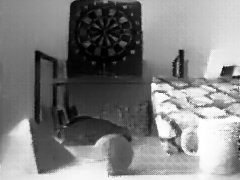}
    & \includegraphics[width=\widthplot,height=\heightplot]{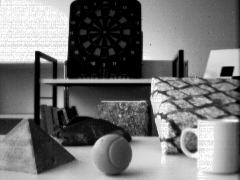}\\[\vspacecols]
    \includegraphics[width=\widthplot,height=\heightplot]{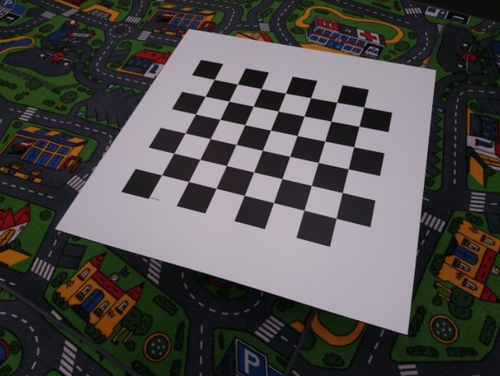}
    & \includegraphics[width=\widthplot,height=\heightplot]{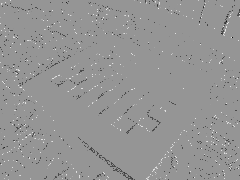}
    & \includegraphics[width=\widthplot,height=\heightplot]{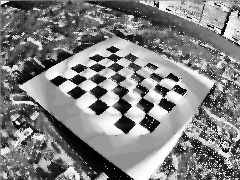}
    & \includegraphics[width=\widthplot,height=\heightplot]{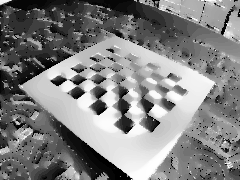}
    & \includegraphics[width=\widthplot,height=\heightplot]{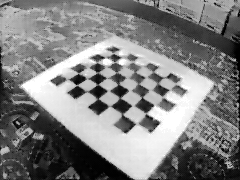}
    & \includegraphics[width=\widthplot,height=\heightplot]{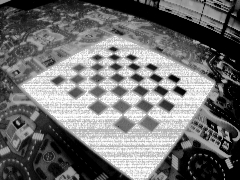}\\[\vspacecols]
    (a) Scene Preview & (b) Events & (c) HF & (d) MR & (e) Ours & (f) Ground truth\\
    \end{tabular}
\vspace{0.5ex}
\caption{Qualitative comparison of our reconstruction method with two recent competing approaches, MR \cite{Munda18ijcv} and HF \cite{Scheerlinck18accv}, on sequences from \cite{Mueggler17ijrr}, which contain ground truth frames from a DAVIS240C sensor. Our method successfully reconstructs fine details (textures in the second and third row) compared to other methods, while avoiding ghosting effects (particulary visible in the shapes sequences on the fourth row).
}
\label{fig:comp_event_camera_dataset_supp}
\end{figure*}

\setlength{\tabcolsep}{0.3ex} %
\global\long\def\heightplot{3.4cm} %
\global\long\def\widthplot{3.4cm} %
\global\long\def\vspacecols{0.3ex} %
\begin{figure*}
	\centering
    \begin{tabular}{ccccc}
    \includegraphics[width=\widthplot,height=\heightplot]{images/comp_Bardow_dataset/events/1_jumping.png}
    & \includegraphics[width=\widthplot,height=\heightplot]{images/comp_Bardow_dataset/comp_MR/jumping/1_SOFIE.png}
    & \includegraphics[width=\widthplot,height=\heightplot]{images/comp_Bardow_dataset/comp_MR/jumping/1_CF_filtered.png}
    & \includegraphics[width=\widthplot,height=\heightplot]{images/comp_Bardow_dataset/comp_MR/jumping/1_MR.png}
    & \includegraphics[width=\widthplot,height=\heightplot]{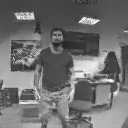}\\[\vspacecols]
    \includegraphics[width=\widthplot,height=\heightplot]{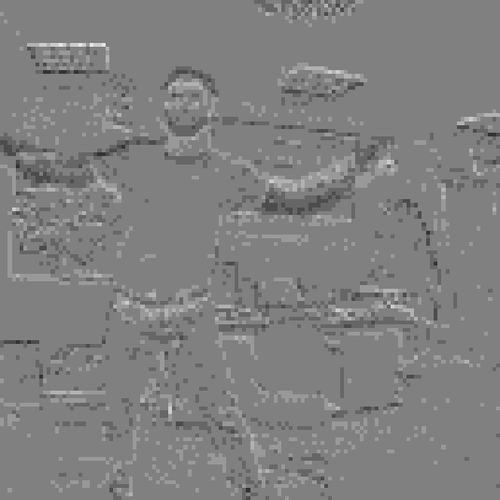}
    & \includegraphics[width=\widthplot,height=\heightplot]{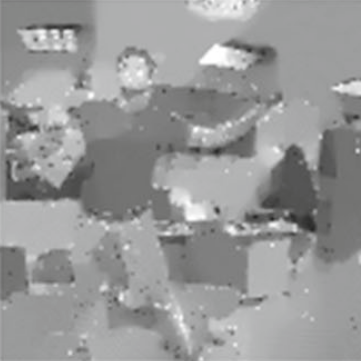}
    & \includegraphics[width=\widthplot,height=\heightplot]{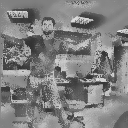}
    & \includegraphics[width=\widthplot,height=\heightplot]{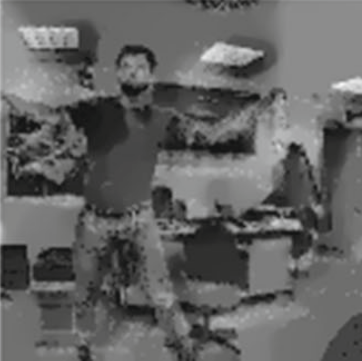}
    & \includegraphics[width=\widthplot,height=\heightplot]{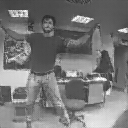}\\[\vspacecols]
    \includegraphics[width=\widthplot,height=\heightplot]{images/comp_Bardow_dataset/events/1_face.png}
    & \includegraphics[width=\widthplot,height=\heightplot]{images/comp_Bardow_dataset/comp_MR/face/1_SOFIE.png}
    & \includegraphics[width=\widthplot,height=\heightplot]{images/comp_Bardow_dataset/comp_MR/face/1_CF_filtered.png}
    & \includegraphics[width=\widthplot,height=\heightplot]{images/comp_Bardow_dataset/comp_MR/face/1_MR.png}
    & \includegraphics[width=\widthplot,height=\heightplot]{images/comp_Bardow_dataset/comp_MR/face/1_ours.png}\\[\vspacecols]
    \includegraphics[width=\widthplot,height=\heightplot]{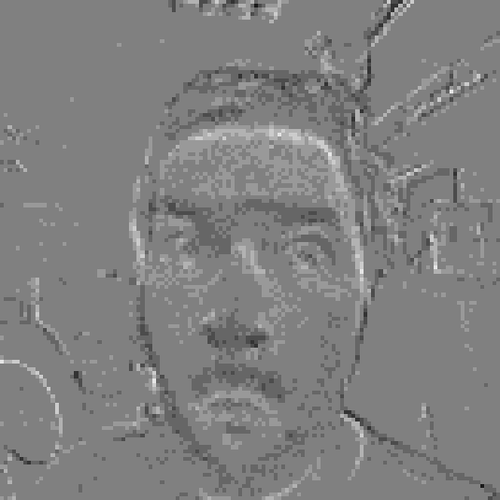}
    & \includegraphics[width=\widthplot,height=\heightplot]{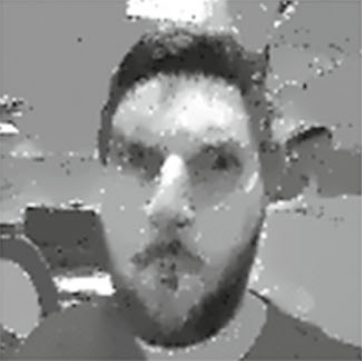}
    & \includegraphics[width=\widthplot,height=\heightplot]{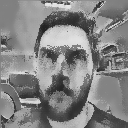}
    & \includegraphics[width=\widthplot,height=\heightplot]{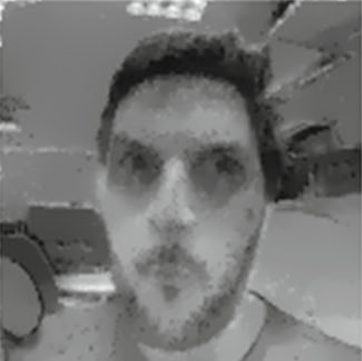}
    & \includegraphics[width=\widthplot,height=\heightplot]{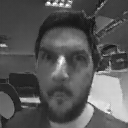}\\[\vspacecols]
    \includegraphics[width=\widthplot,height=\heightplot]{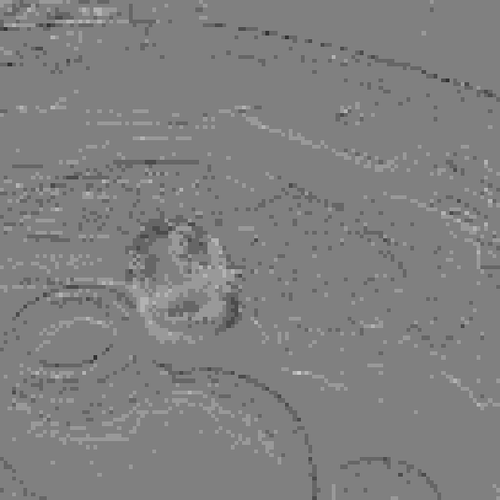}
    & \includegraphics[width=\widthplot,height=\heightplot]{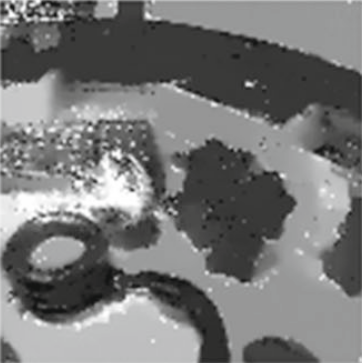}
    & \includegraphics[width=\widthplot,height=\heightplot]{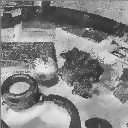}
    & \includegraphics[width=\widthplot,height=\heightplot]{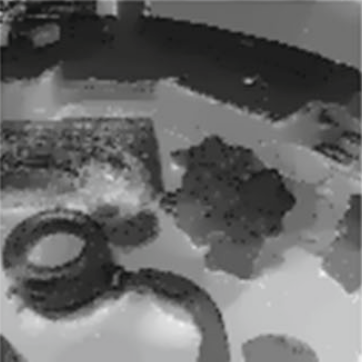}
    & \includegraphics[width=\widthplot,height=\heightplot]{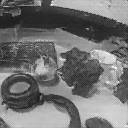}\\[\vspacecols]
    (a) Events & (b) SOFIE \cite{Bardow16cvpr} & (c) HF \cite{Scheerlinck18accv} & (d) MR \cite{Munda18ijcv} & (e) Ours\\
    \end{tabular}
\vspace{0.5ex}
\caption{Qualitative comparison of our reconstruction method with various competing approaches. We used the datasets from \cite{Bardow16cvpr}. The dataset does not contain ground truth images, thus only a qualitative comparison is possible. For SOFIE and MR, we used images provided by the authors, for which the parameters were tuned for each dataset. For HF, we ran the code provided by the authors, manually tuned the parameters on these datasets to achieve the best visual quality, and additionally applied a bilateral filter to clean the high frequency noise present in the original reconstructions.
}
\label{fig:comp_bardow_supp}
\vspace{6ex}
\end{figure*}

\global\long\def\heightplot{3.5cm} %
\global\long\def\widthplot{4.658cm} %
\begin{figure*}
	\centering
    \begin{tabular}{ccc}
    \includegraphics[width=\widthplot,height=\heightplot]{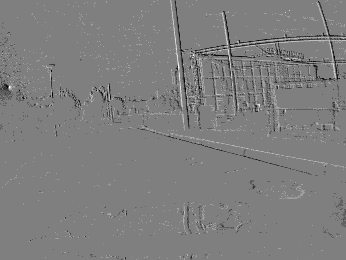}
    & \includegraphics[width=\widthplot,height=\heightplot]{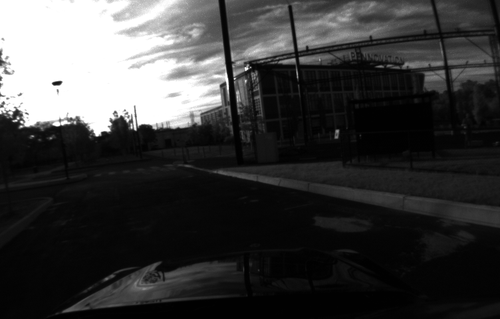}
    & \includegraphics[width=\widthplot,height=\heightplot]{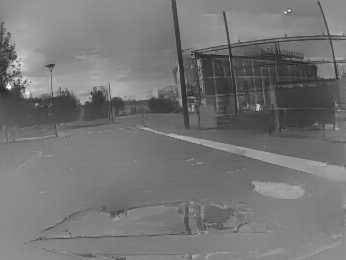}\\[\vspacecols]
    \includegraphics[width=\widthplot,height=\heightplot]{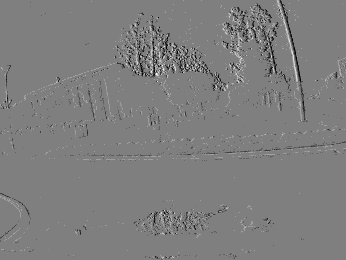}
    & \includegraphics[width=\widthplot,height=\heightplot]{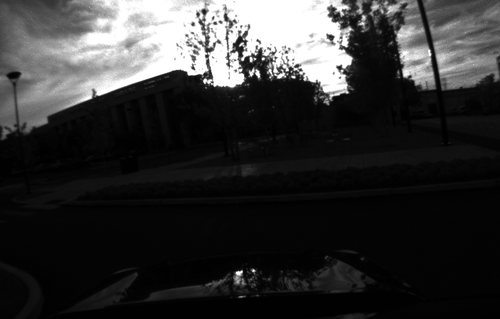}
    & \includegraphics[width=\widthplot,height=\heightplot]{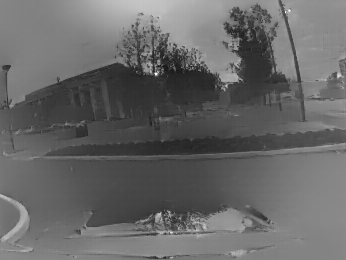}\\[\vspacecols]
    \includegraphics[width=\widthplot,height=\heightplot]{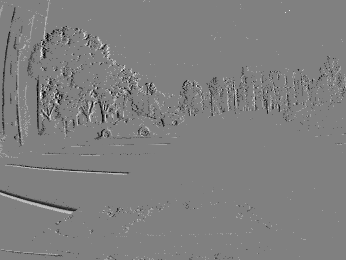}
    & \includegraphics[width=\widthplot,height=\heightplot]{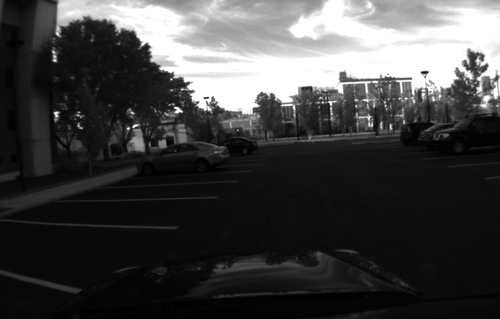}
    & \includegraphics[width=\widthplot,height=\heightplot]{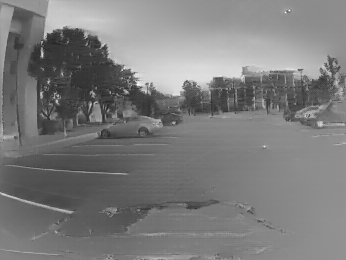}\\[\vspacecols]
    \includegraphics[width=\widthplot,height=\heightplot]{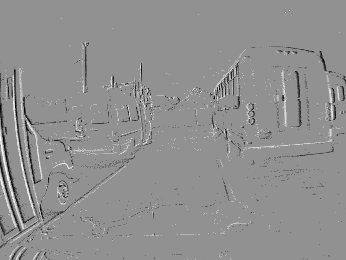}
    & \includegraphics[width=\widthplot,height=\heightplot]{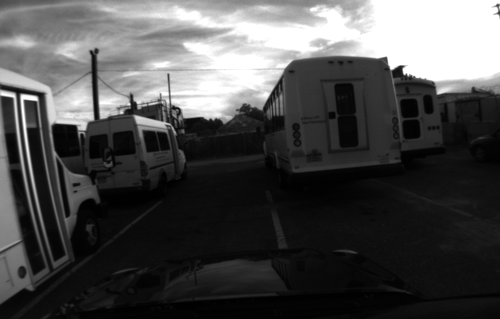}
    & \includegraphics[width=\widthplot,height=\heightplot]{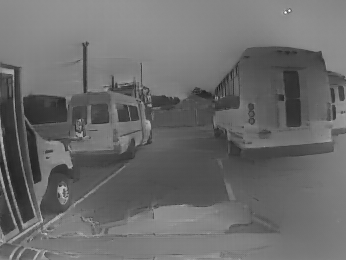}\\[\vspacecols]
    \includegraphics[width=\widthplot,height=\heightplot]{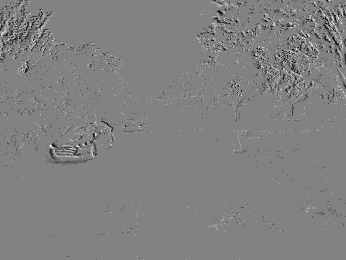}
    & \includegraphics[width=\widthplot,height=\heightplot]{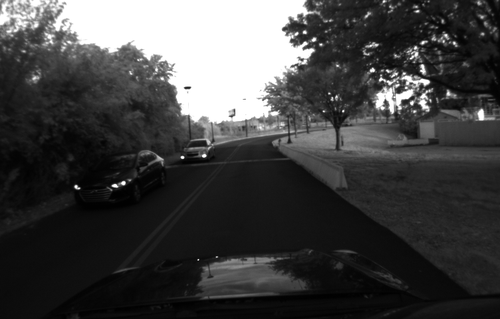}
    & \includegraphics[width=\widthplot,height=\heightplot]{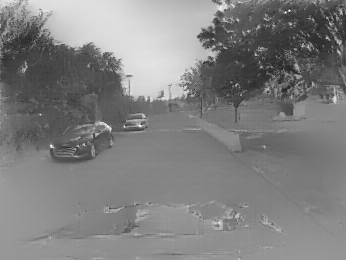}\\[\vspacecols]
    (a) Events & (b) VI sensor frame & (c) Our reconstruction\\
    \end{tabular}
\vspace{1ex}
\caption{Example HDR reconstructions on the MVSEC automotive dataset \cite{Zhu18ral}. The standard frames were recorded with a high-quality VI sensor with auto-exposure activated. Because the camera is facing directly the sun, the standard frames (b) are either under- or over-exposed since the limited dynamic range of the standard sensor cannot cope with the high dynamic range of the scene. By contrast, the events (a) capture the whole dynamic range of the scene, which our method successfully reconstructs to high dynamic range images (c), allow to discover details that were not visible in the standard frames.
}
\label{fig:qualitative_hdr_supp}
\vspace{4ex}
\end{figure*}

\global\long\def\heightplot{3.5cm} %
\global\long\def\widthplot{4.658cm} %
\begin{figure*}
	\centering
    \begin{tabular}{ccc}
    \includegraphics[width=\widthplot,height=\heightplot]{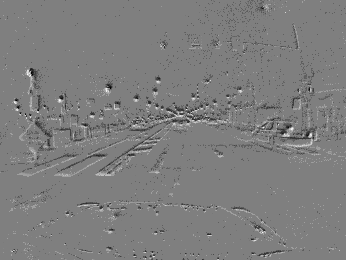}
    & \includegraphics[width=\widthplot,height=\heightplot]{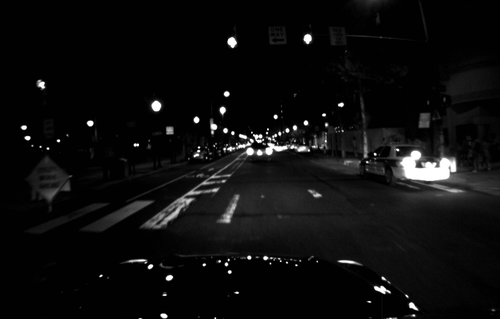}
    & \includegraphics[width=\widthplot,height=\heightplot]{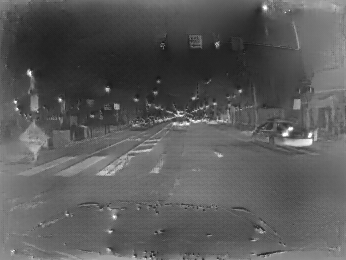}\\[\vspacecols]
    \includegraphics[width=\widthplot,height=\heightplot]{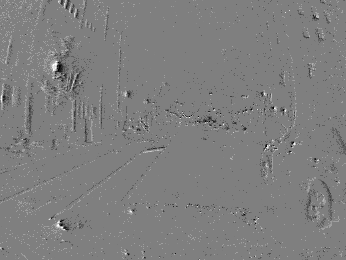}
    & \includegraphics[width=\widthplot,height=\heightplot]{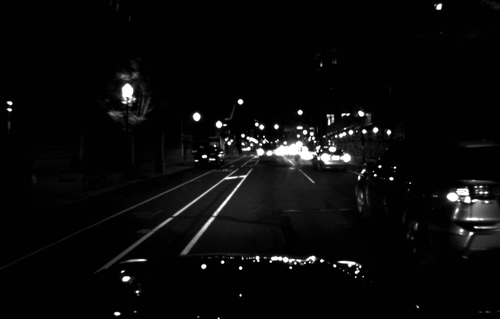}
    & \includegraphics[width=\widthplot,height=\heightplot]{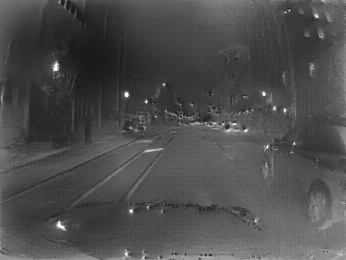}\\[\vspacecols]
    \includegraphics[width=\widthplot,height=\heightplot]{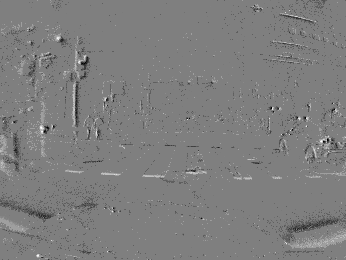}
    & \includegraphics[width=\widthplot,height=\heightplot]{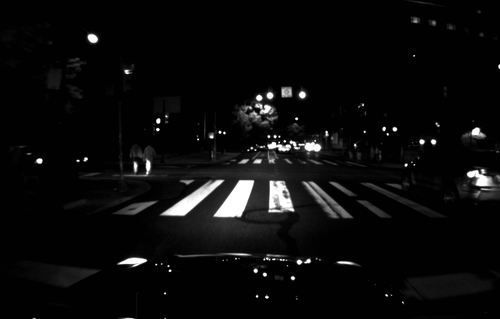}
    & \includegraphics[width=\widthplot,height=\heightplot]{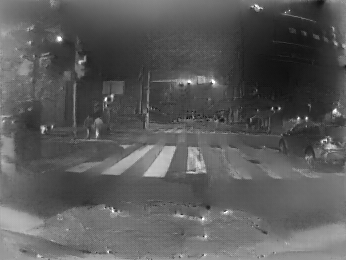}\\[\vspacecols]
    \includegraphics[width=\widthplot,height=\heightplot]{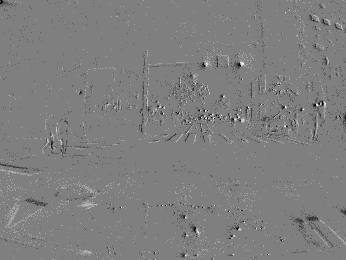}
    & \includegraphics[width=\widthplot,height=\heightplot]{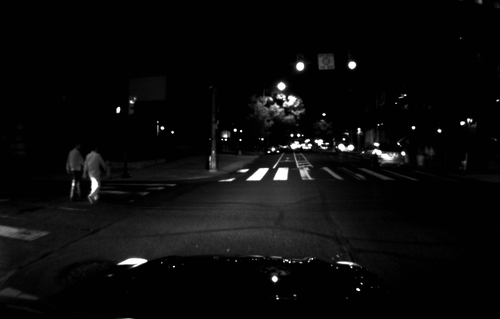}
    & \includegraphics[width=\widthplot,height=\heightplot]{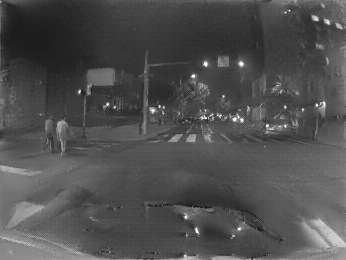}\\[\vspacecols]
    \includegraphics[width=\widthplot,height=\heightplot]{images/qualitative/MVSEC/night/1098_events.png}
    & \includegraphics[width=\widthplot,height=\heightplot]{images/qualitative/MVSEC/night/1098_VISENSOR.png}
    & \includegraphics[width=\widthplot,height=\heightplot]{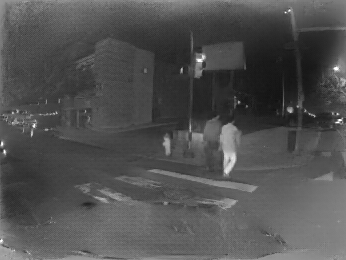}\\[\vspacecols]
    (a) Events & (b) VI sensor frame & (c) Our reconstruction\\
    \end{tabular}
\vspace{1ex}
\caption{Example HDR reconstructions on the MVSEC automotive dataset \cite{Zhu18ral} at night. The standard frames were recorded with a high-quality VI sensor with auto-exposure activated. Because of low light during the night, the standard frames (b) are severely degraded. By contrast, the events (a) still can capture the whole dynamic range of the scene, which our method successfully recovers (c), allowing to discover details that were not visible in the standard frames.
}
\label{fig:qualitative_night_supp}
\vspace{6ex}
\end{figure*}

\section{Object Classification}

Below we detail the exact modalities of our reconstruction method for each of the dataset which we used for our evaluation of object classification (Section 5.1 in the paper), as well as the specific architectures used and training modalities.

\mypara{N-MNIST.}
To reconstruct images with our networks, we used an event window of $\NumEvents=1@000$ events.
We passed every event sequence into our network, resulting in a video, from which we keep the final image as input for the classification network.
To match the images from the original MNIST dataset, we additionally binarize the reconstructed image (whose values lie in $[0,1]$) with a threshold of $0.5$.
The train and test images were normalized so that the mean value of each image is $0.1307$ and the variance $0.3081$.
We used the official train and test split provided in the M-NNIST dataset.
As there is no standard state of the art architecture for MNIST, we used a simple CNN architecture as our classification network, composed of the following blocks:

\begin{itemize}
  \item \small 2D convolution (stride: 5, output channels: 32) + ReLU
  \vspace{-2mm}
  \item \small 2D convolution (stride: 5, output channels: 64) + ReLU
  \vspace{-2mm}
  \item \small 2D max pooling (size: 2) + Dropout
  \vspace{-2mm}
  \item \small Fully connected layer (output size: 128 neurons) + ReLU
  \vspace{-2mm}
  \item \small Fully connected layer (output size: 10 neurons)
\end{itemize}

We used the cross entropy loss, and trained the network for 15 epochs using the ADAM optimizer, with a learning rate of 0.001.

\mypara{N-CARS.}
We used windows of events with a fixed temporal size of $\SI{20}{\ms}$, and used the last reconstructed image from the video as input to the classification network.
We used the official train and test split provided by the N-CARS dataset.
We used a ResNet18 \cite{He16cvpr} architecture (with an additional fully connected final layer with $2$ output neurons), initialized with weights pretrained on ImageNet \cite{Russakovsky15ijcv}, and fine-tuned the network using the reconstructed images from the training set for 20 epochs, using SGD with a learning rate of $0.001$ (decayed by factor of $0.1$ every $7$ epochs), and momentum of $0.1$.

\mypara{N-Caltech101.}
For image reconstruction, we used windows of $\NumEvents=10@000@$ events and used the last reconstructed image as input to the classification network.
Since there is no official train and test split for the N-Caltech101 dataset, we split the dataset randomly into two third training sequences ($5@863$ sequences) and one third testing sequences ($2@396$ sequences), following the methodology used by HATS \cite{Sironi18cvpr}.
The train and test images were converted to 3-channel grayscale images (\ie the three channels are the same), and normalized so that the mean value of each image is $0.485$ and the variance $0.229$.
We also performed data augmentation at train time (random horizontal flips, and random crop of size $224$).
At test time, we resized all the images to $256 \timess 256$ and cropped the image around the center with a size of $224$.
We used a ResNet18 architecture (with an additional fully-connected final layer with $101$ output neurons), initialized with weights pretrained on ImageNet, and fine-tuned the network using the reconstructed images from the training set for 25 epochs using SGD with an initial learning rate of $0.001$ (decayed by a factor of $0.1$ every 7 epochs) and momentum of $0.1$.
Fig.~\ref{fig:preview_ncaltech101_supp} shows additional reconstruction examples from the N-Caltech101 dataset.

\setlength{\tabcolsep}{0.3ex} %
\global\long\def\heightplot{2.8cm} %
\global\long\def\vspacecols{0.3ex} %
\begin{figure*}
	\centering
    \begin{tabular}{ccc}
    \includegraphics[height=\heightplot]{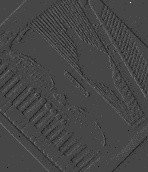}
    & \includegraphics[height=\heightplot]{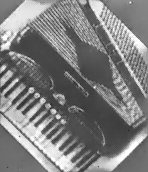}
    & \includegraphics[height=\heightplot]{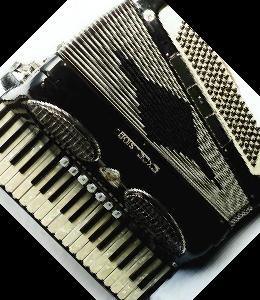}\\[\vspacecols]
    \includegraphics[height=\heightplot]{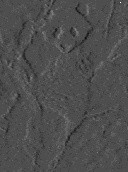}
    & \includegraphics[height=\heightplot]{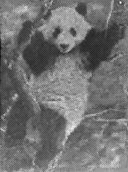}
    & \includegraphics[height=\heightplot]{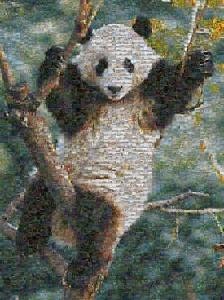}\\[\vspacecols]
    \includegraphics[height=\heightplot]{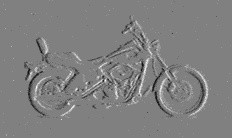}
    & \includegraphics[height=\heightplot]{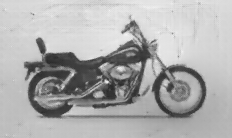}
    & \includegraphics[height=\heightplot]{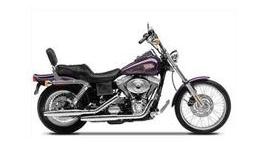}\\[\vspacecols]
    \includegraphics[height=\heightplot]{images/N-Caltech101/wild_cat/events_0001.jpg}
    & \includegraphics[height=\heightplot]{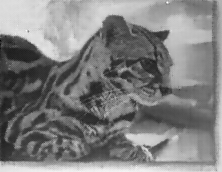}
    & \includegraphics[height=\heightplot]{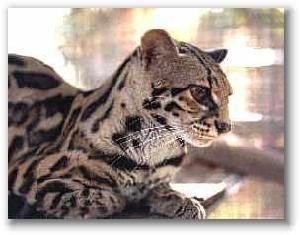}\\[\vspacecols]
    \includegraphics[height=\heightplot]{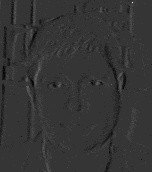}
    & \includegraphics[height=\heightplot]{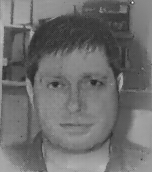}
    & \includegraphics[height=\heightplot]{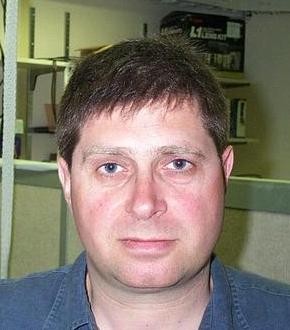}\\[\vspacecols]
    \includegraphics[height=\heightplot]{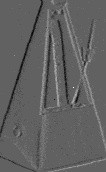}
    & \includegraphics[height=\heightplot]{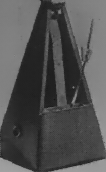}
    & \includegraphics[height=\heightplot]{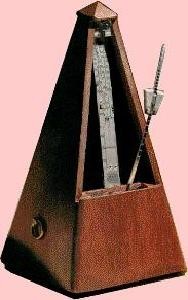}\\[\vspacecols]
    \includegraphics[height=\heightplot]{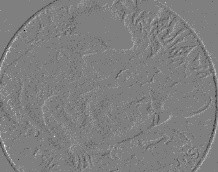}
    & \includegraphics[height=\heightplot]{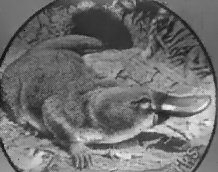}
    & \includegraphics[height=\heightplot]{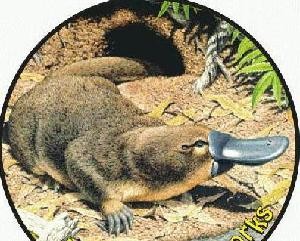}\\[\vspacecols]
    (a) Events & (b) Our Reconstruction & (c) Original Image\\
    \end{tabular}
\vspace{0.5ex}
\caption{(a) Previews of some event sequences from the N-Caltech101 dataset \cite{Orchard15fns} which features event sequences converted from the Caltech101 dataset. (b) our reconstructions (from events only) preserve many of the details and statistics of the original images (c). Note that these datasets feature planar motion (since Caltech101 images were projected on white wall to record the events), which coincides with the type of motions present in the simulated data, which explains in part the outstanding visual quality of the reconstructions.
}
\label{fig:preview_ncaltech101_supp}
\end{figure*}

\section{Visual-Inertial Odometry}

Figs.~\ref{fig:boxplots_supp_overall},~\ref{fig:boxplots_supp_translation} and~\ref{fig:boxplots_supp_6dof} provide additional results on the visual-inertial odometry experiments presented in the main paper.
Specifically, they provide, for each sequence used in our evaluation, the evolution of the mean translation and rotation error as a function of the travelled distance for our approach, UltimateSLAM (E+I), and UltimateSLAM (E+F+I).

\setlength{\tabcolsep}{0.1ex} %
\global\long\def\heightplot{2.9cm} %
\global\long\def\vspacecols{0.0ex} %
\begin{figure*}
	\centering
    \begin{tabular}{cc}
    \includegraphics[height=\heightplot,trim={0 0 8cm 0},clip,]{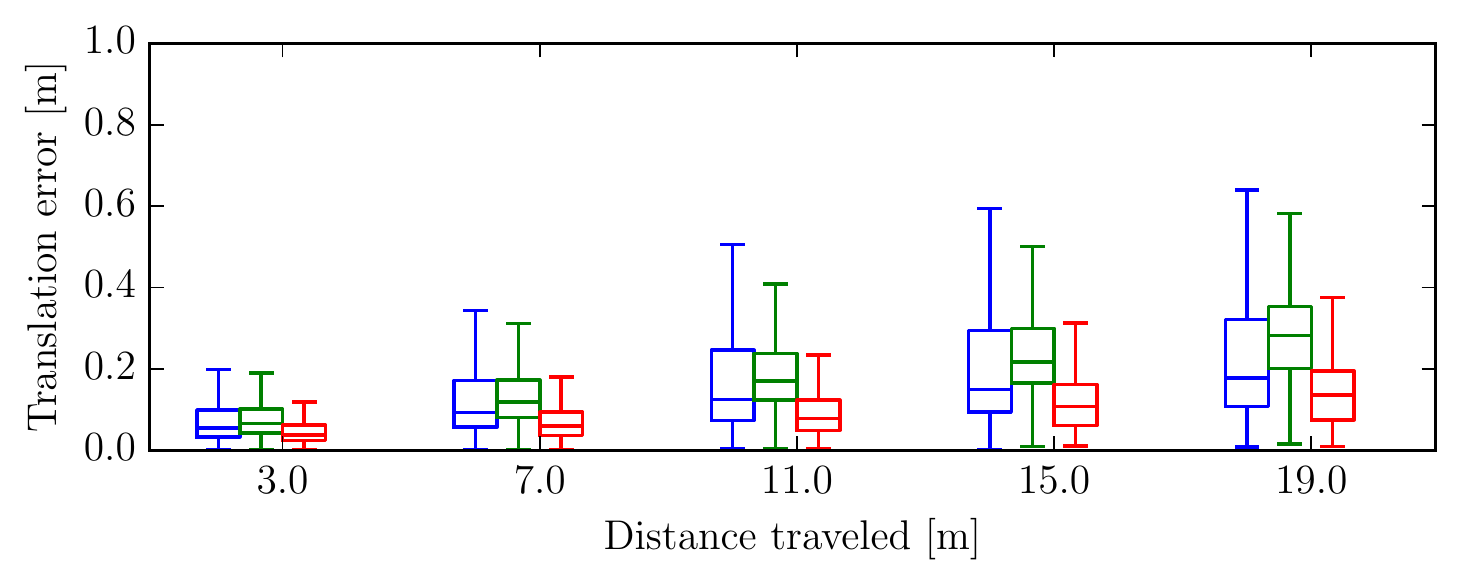}
    & \includegraphics[height=\heightplot]{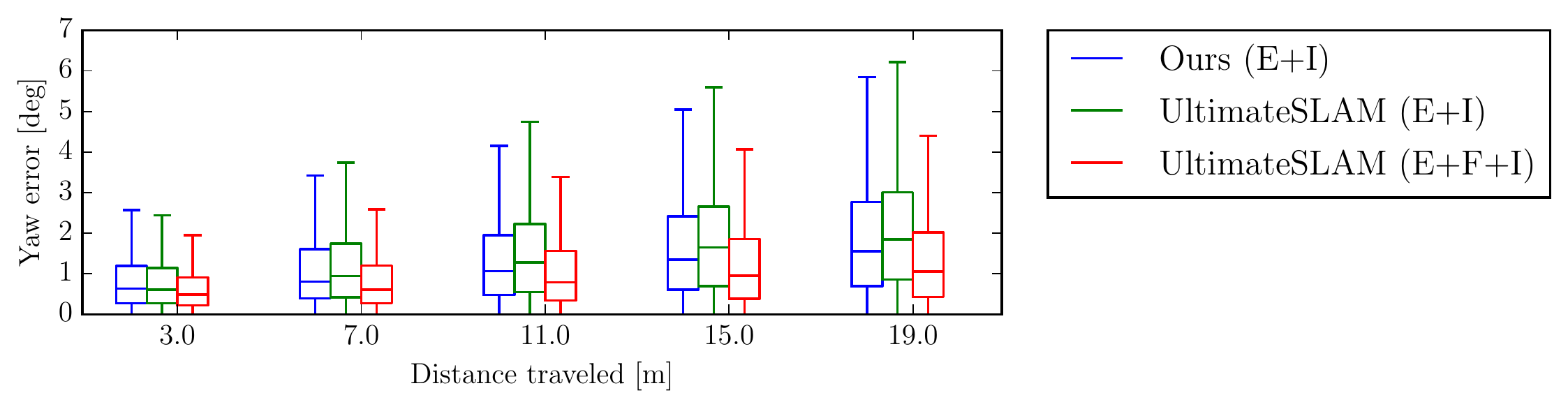}\\[0mm]
    \end{tabular}
\vspace{0.5ex}
\caption{Evolution of the overall mean translation error (in meters) and mean rotation error (in degrees), averaged across all the datasets used in our evaluation.}
\label{fig:boxplots_supp_overall}
\end{figure*}

\setlength{\tabcolsep}{0.1ex} %
\global\long\def\heightplot{2.9cm} %
\global\long\def\vspacecols{0.0ex} %
\begin{figure*}
	\centering
    \begin{tabular}{cc}
    \includegraphics[height=\heightplot,trim={0 0 8cm 0},clip,]{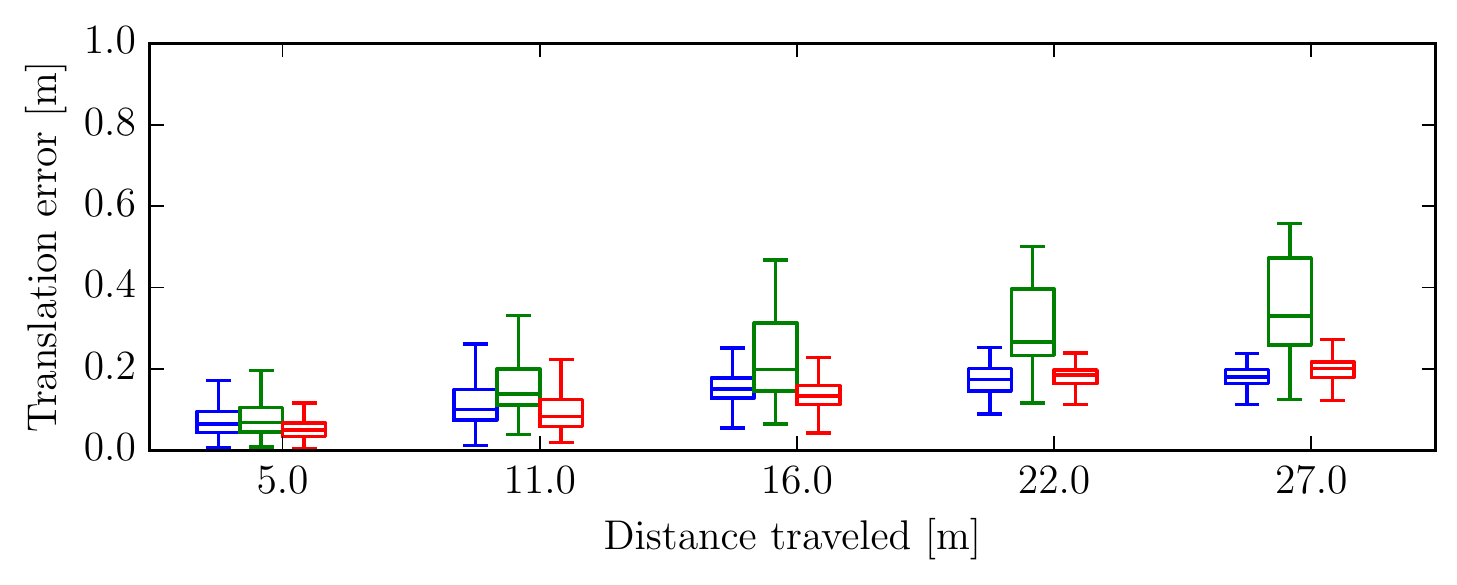}
    & \includegraphics[height=\heightplot]{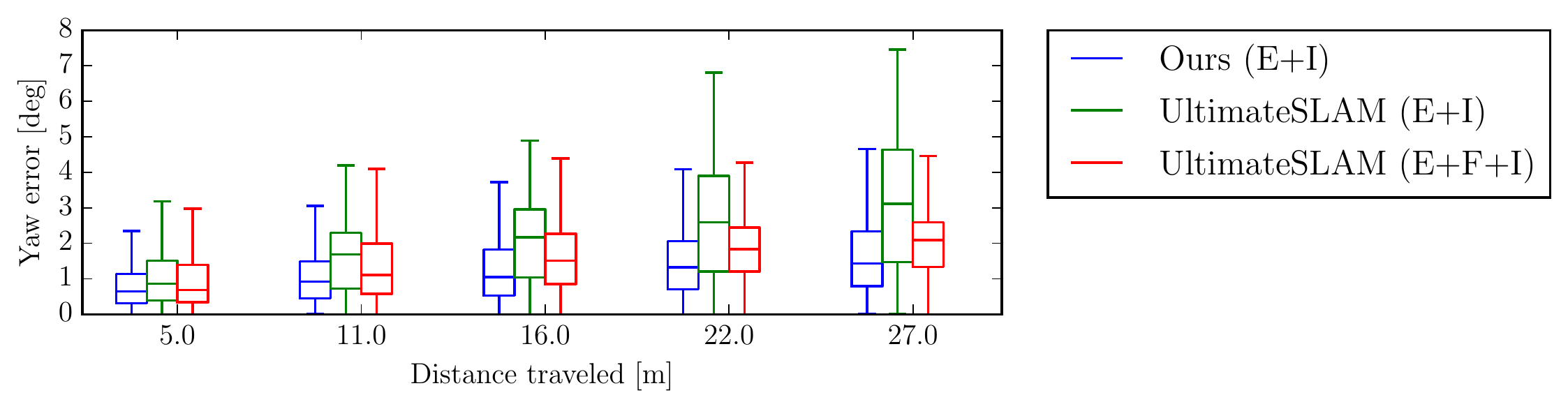}\\[0mm]
    \includegraphics[height=\heightplot,trim={0 0 8cm 0},clip,]{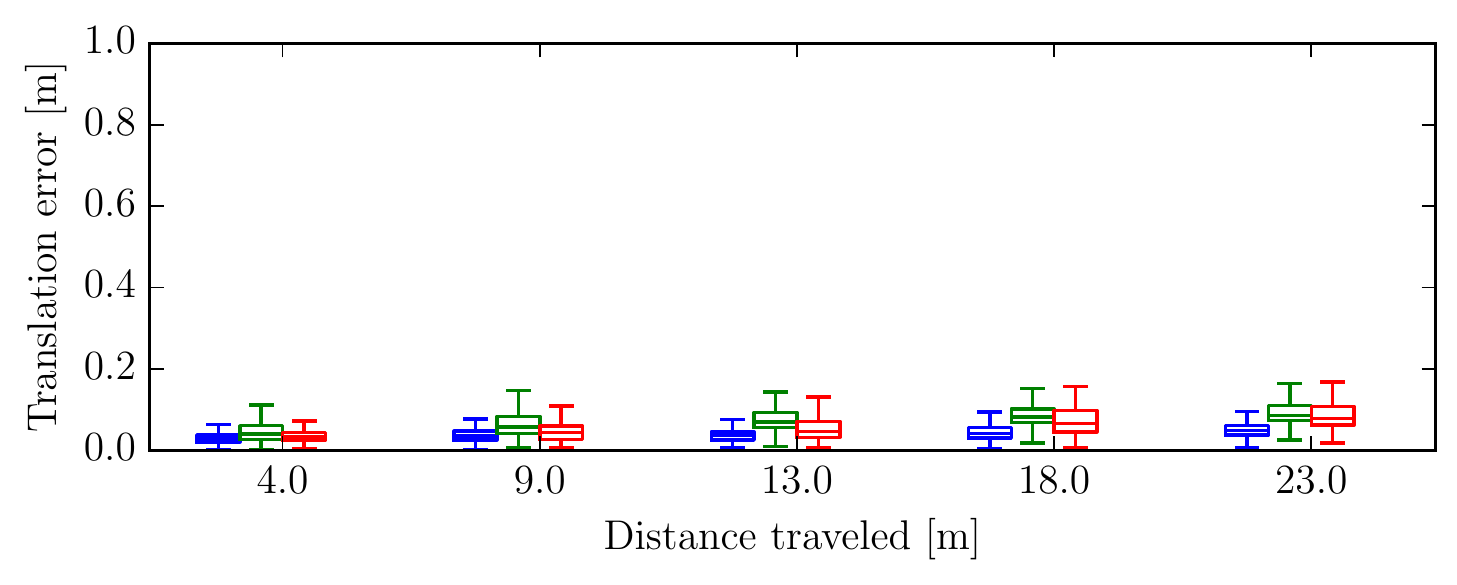}
    & \includegraphics[height=\heightplot]{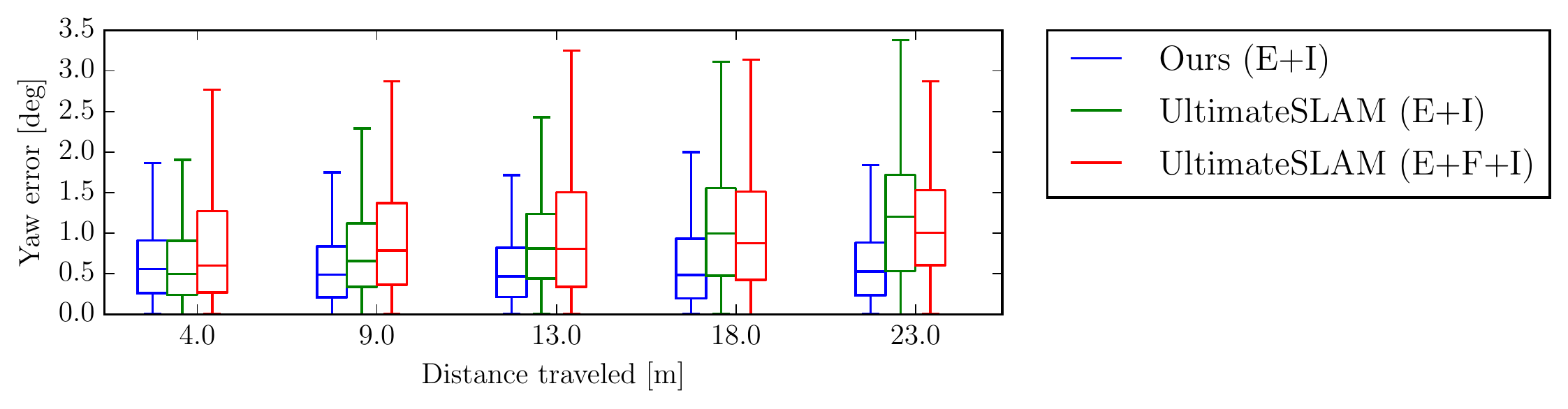}\\[0mm]
    \includegraphics[height=\heightplot,trim={0 0 8cm 0},clip,]{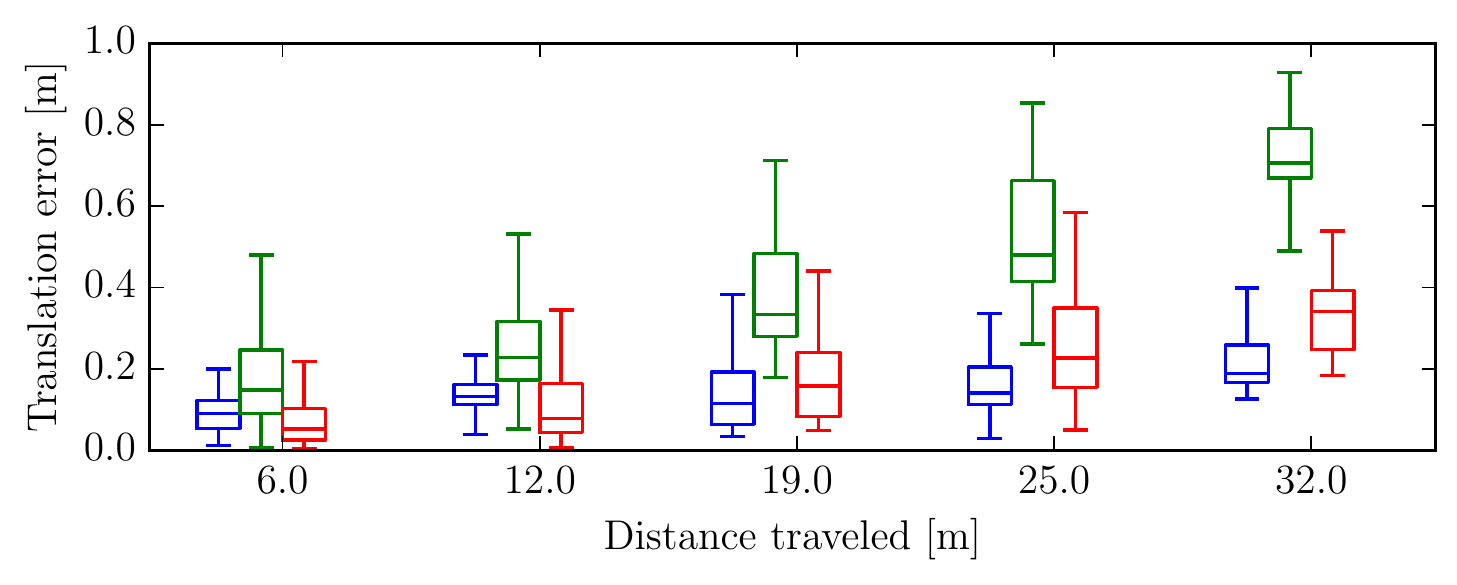}
    & \includegraphics[height=\heightplot]{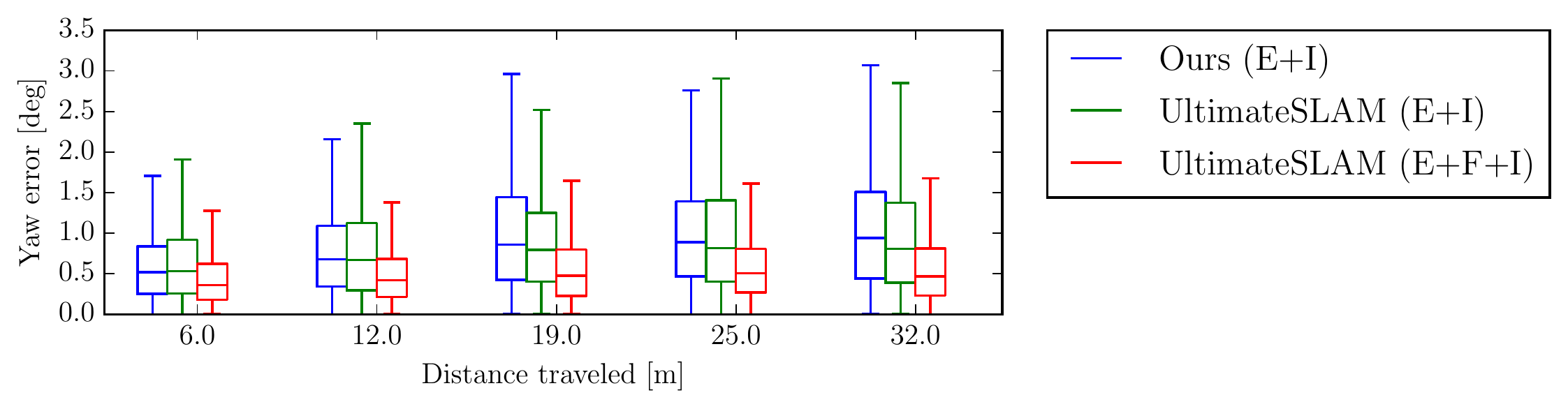}\\[0mm]
    \includegraphics[height=\heightplot,trim={0 0 8cm 0},clip,]{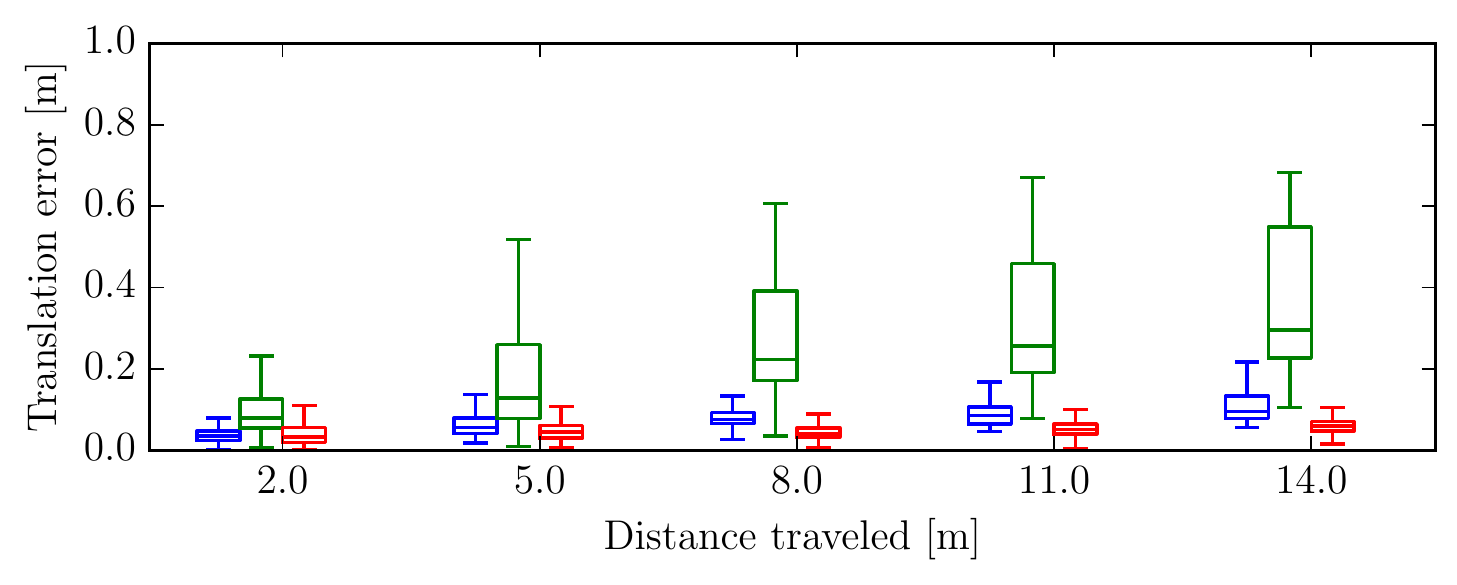}
    & \includegraphics[height=\heightplot]{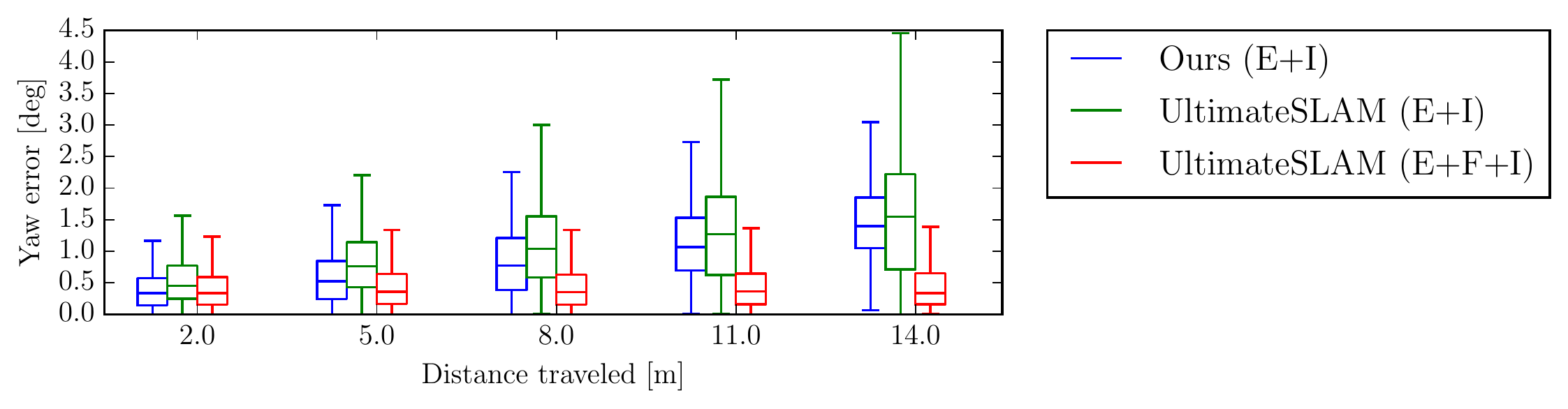}\\[0mm]
    \end{tabular}
\vspace{0.5ex}
\caption{Evolution of the mean translation error (in meters) and mean rotation error (in degrees), as a function of the travelled distance. Sequences from top to bottom: 'shapes\_translation', 'poster\_translation', 'boxes\_translation', 'dynamic\_translation'.}
\label{fig:boxplots_supp_translation}
\end{figure*}

\setlength{\tabcolsep}{0.1ex} %
\global\long\def\heightplot{2.9cm} %
\global\long\def\vspacecols{0.0ex} %
\begin{figure*}
	\centering
    \begin{tabular}{cc}
    \includegraphics[height=\heightplot,trim={0 0 8cm 0},clip,]{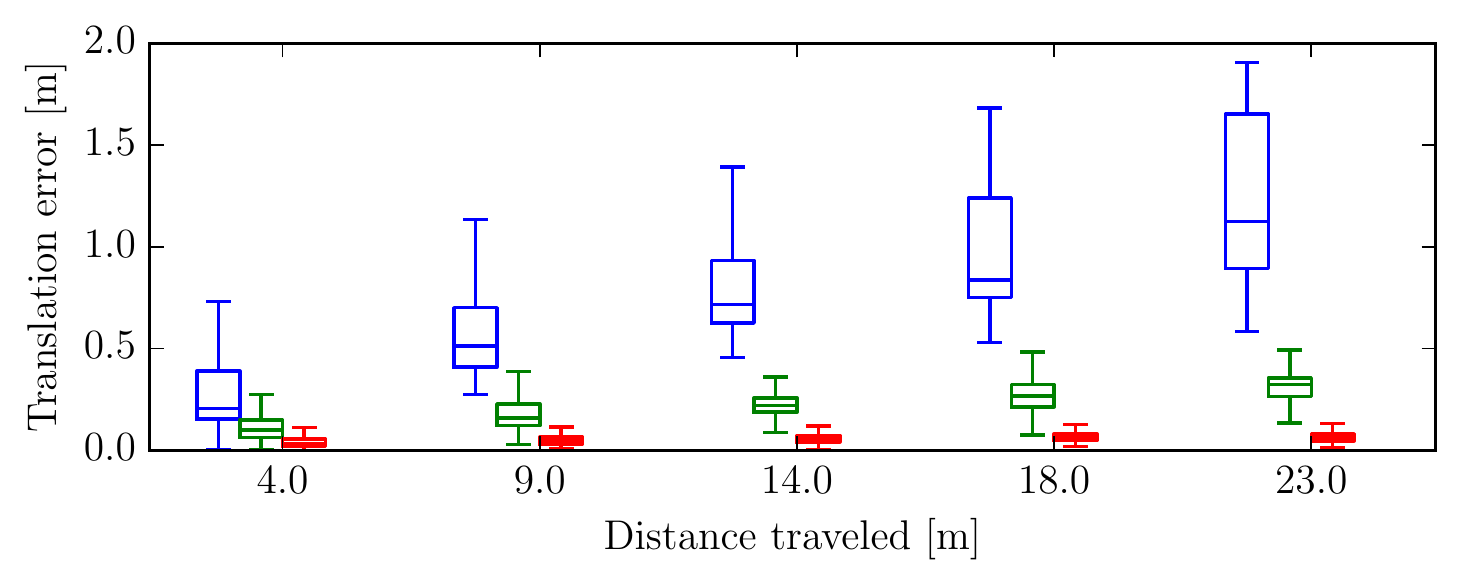}
    & \includegraphics[height=\heightplot]{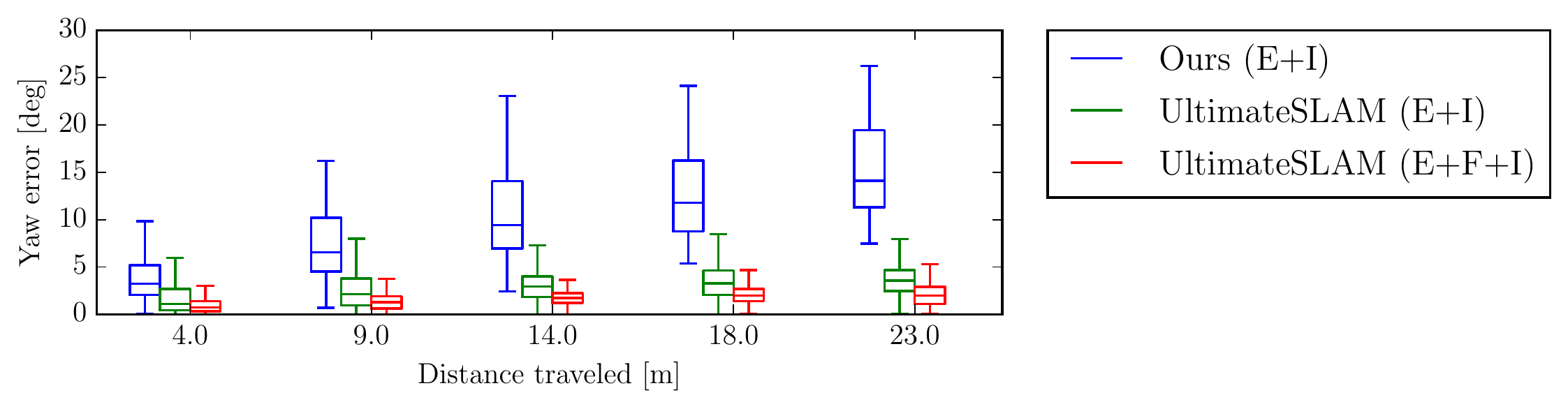}\\[0mm]
    \includegraphics[height=\heightplot,trim={0 0 8cm 0},clip,]{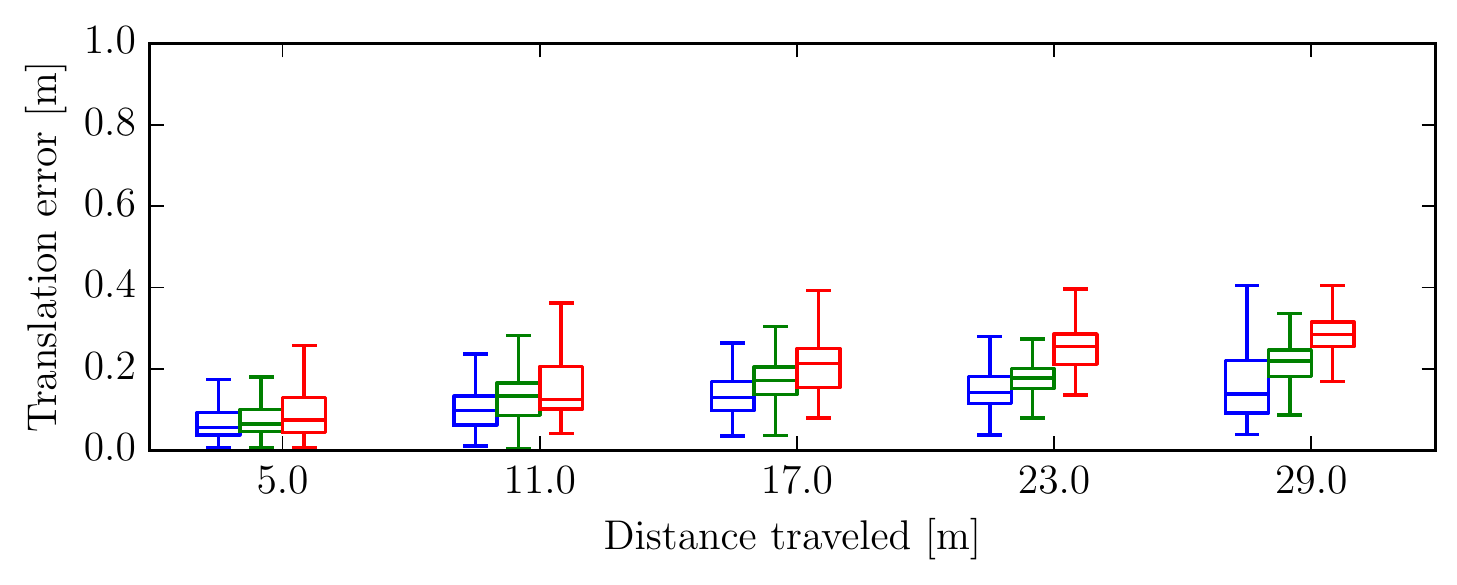}
    & \includegraphics[height=\heightplot]{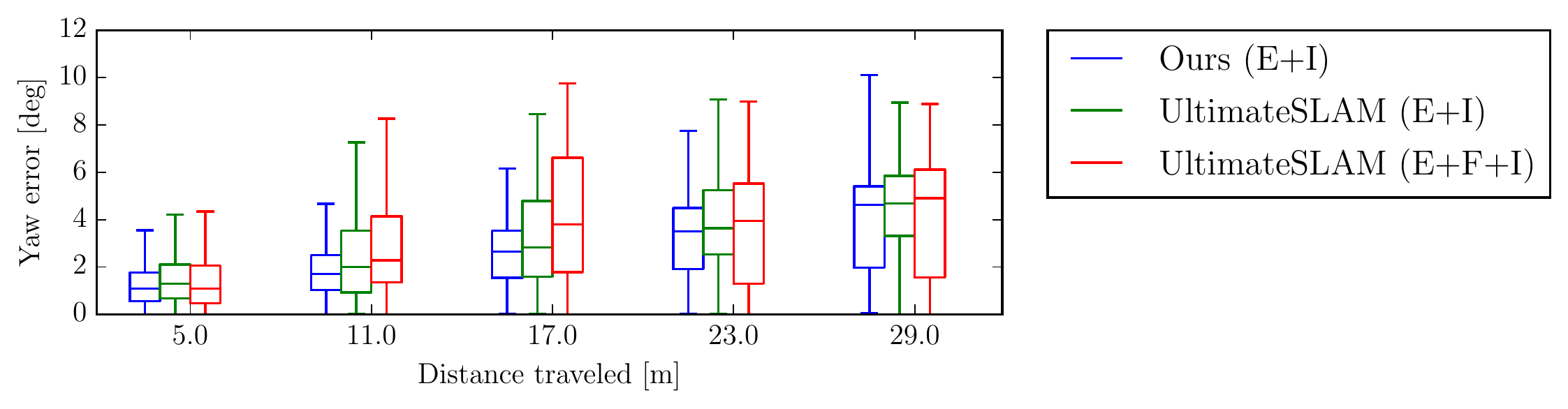}\\[0mm]
    \includegraphics[height=\heightplot,trim={0 0 8cm 0},clip,]{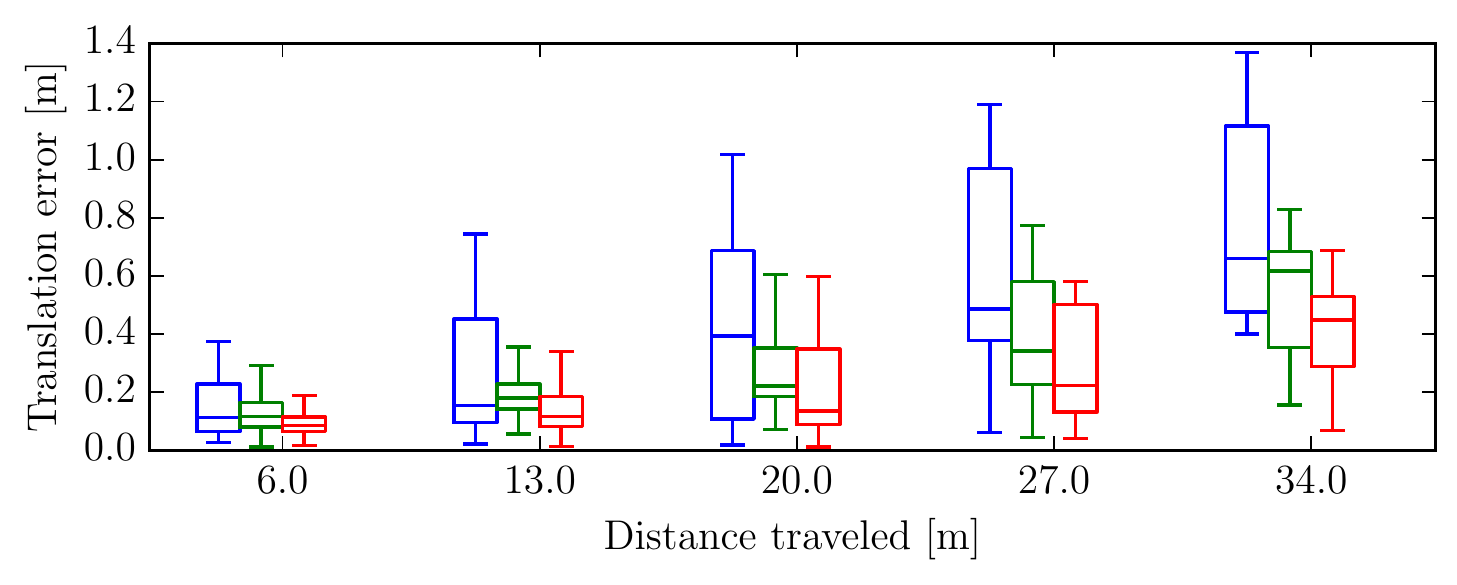}
    & \includegraphics[height=\heightplot]{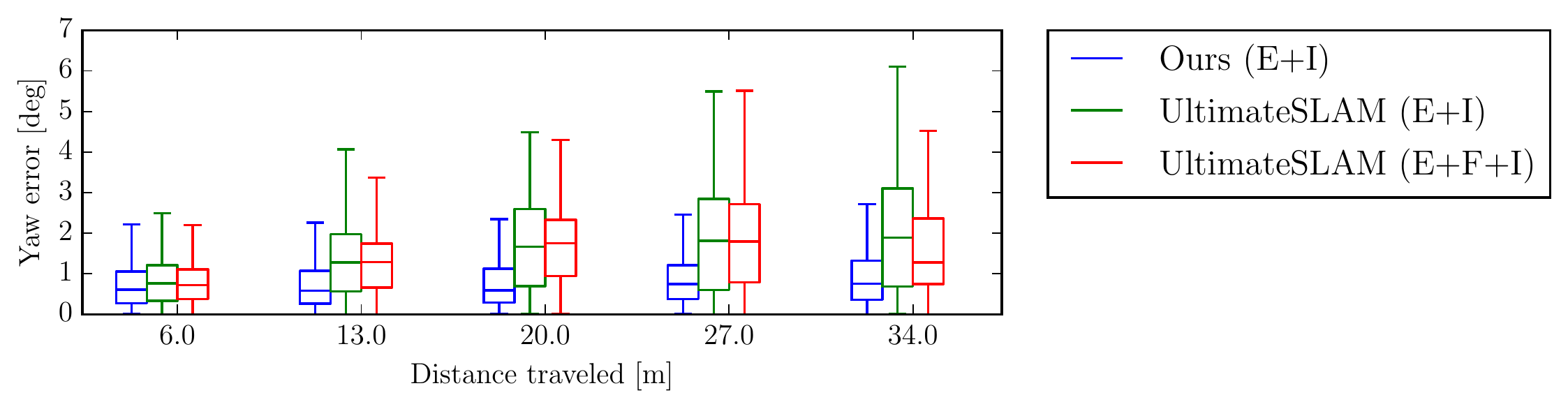}\\
    \includegraphics[height=\heightplot,trim={0 0 8cm 0},clip,]{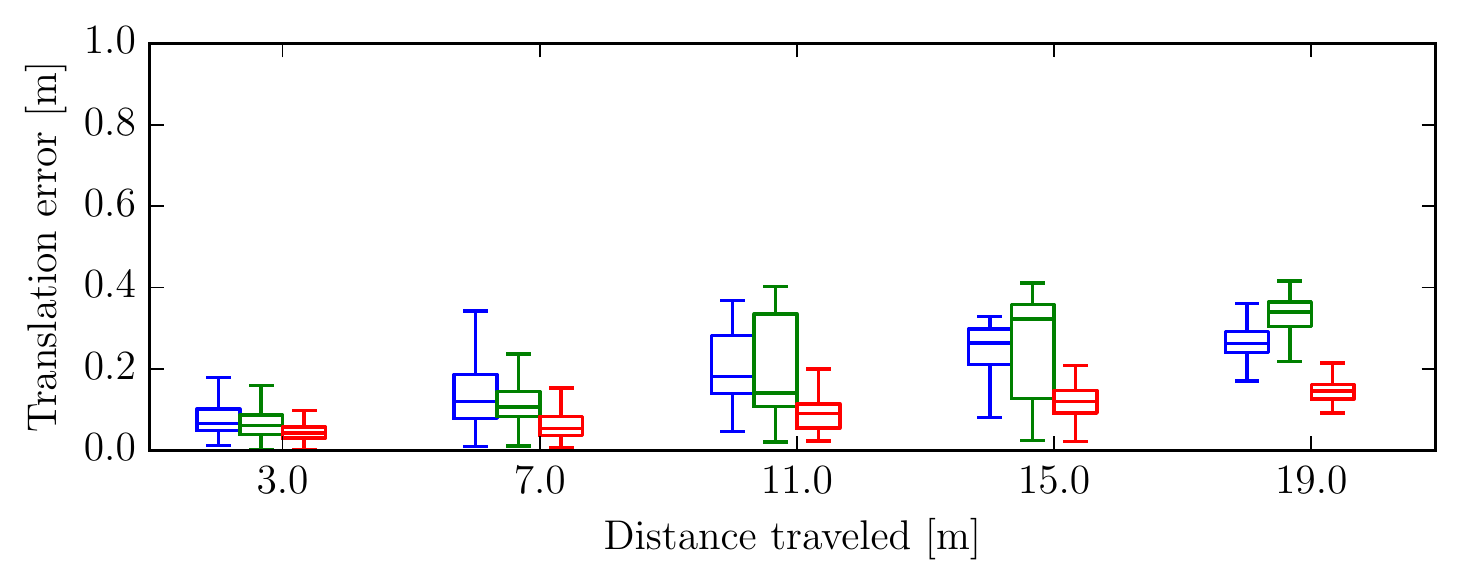}
    & \includegraphics[height=\heightplot]{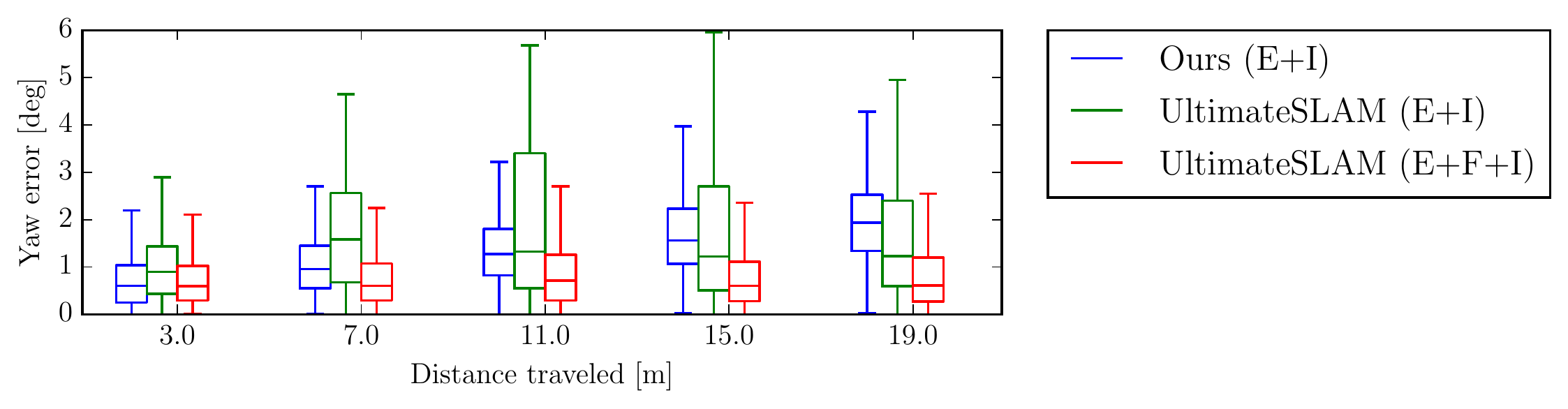}\\[0mm]
    \includegraphics[height=\heightplot,trim={0 0 8cm 0},clip,]{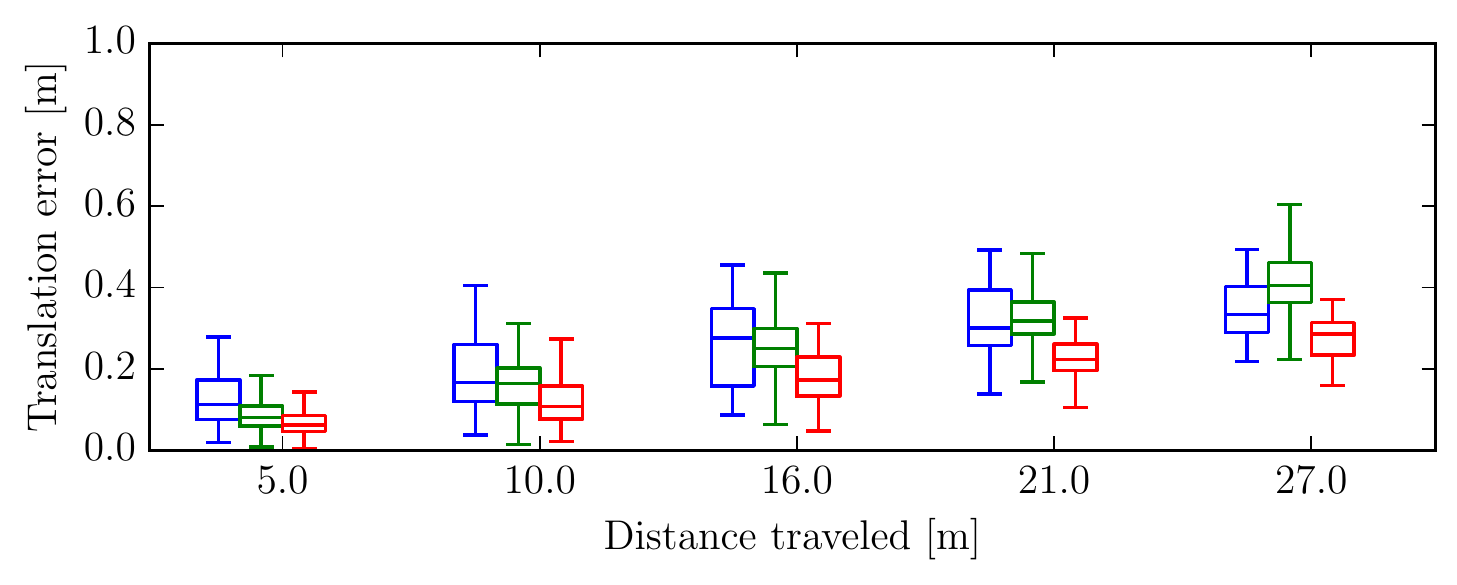}
    & \includegraphics[height=\heightplot]{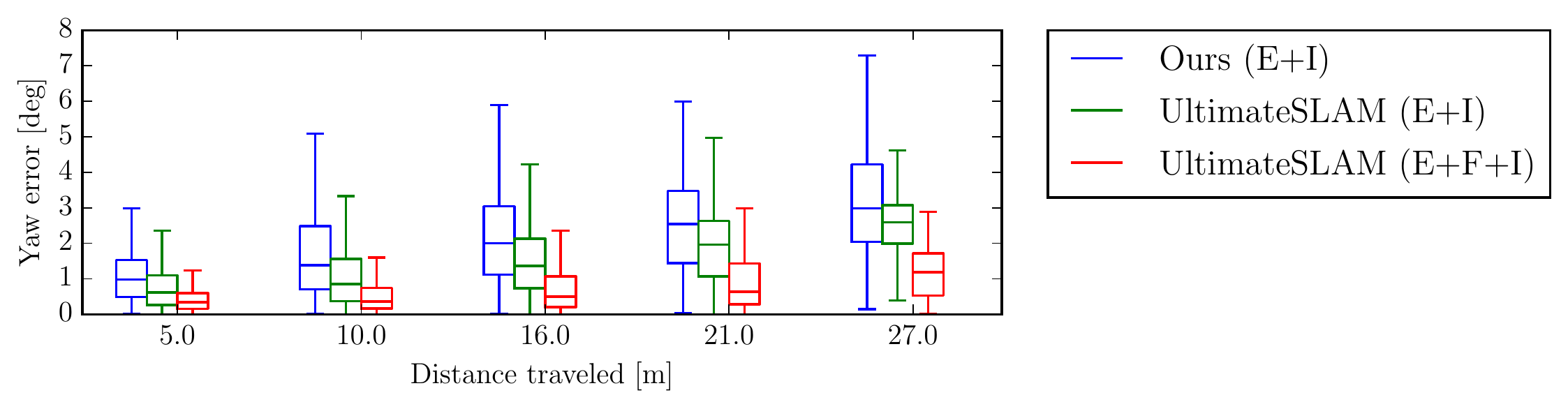}\\[0mm]

    \end{tabular}
\vspace{0.5ex}
\caption{Evolution of the mean translation error (in meters) and mean rotation error (in degrees), as a function of the travelled distance. Sequences from top to bottom: 'shapes\_6dof', 'poster\_6dof', 'boxes\_6dof', 'dynamic\_6dof', 'hdr\_boxes'.}
\label{fig:boxplots_supp_6dof}
\end{figure*}